\PassOptionsToPackage{table}{xcolor}
\documentclass[sigconf]{acmart} 
  
\usepackage{tabularx}
\usepackage{placeins}
\usepackage[most]{tcolorbox}
\usepackage{fvextra}
\usepackage{enumitem}

\renewcommand\footnotetextcopyrightpermission[1]{}

\title{EmbGen: Teaching with Reassembled Corpora}

\author{Arun Lenin}
\affiliation{
  \institution{Commonwealth Bank of Australia}
  \city{Sydney}
  \country{Australia}
}
\email{arun.lenin@cba.com.au}

\author{Kai Rouse}
\affiliation{
  \institution{Commonwealth Bank of Australia}
  \city{Sydney}
  \country{Australia}
}
\email{kai.rouse@cba.com.au}

\author{Andrea Nicastro}
\affiliation{
  \institution{Commonwealth Bank of Australia}
  \city{Sydney}
  \country{Australia}
}
\email{andrea.nicastro@cba.com.au}

\author{Anna Leontjeva}
\affiliation{
  \institution{Commonwealth Bank of Australia}
  \city{Sydney}
  \country{Australia}
}
\email{anna.leontjeva@cba.com.au}

\usepackage{booktabs}
\usepackage{makecell}
\renewcommand{\arraystretch}{1.0}  

\begin{document}

\nopagecolor
\pagecolor{white}

\begin{abstract}
Adapting small instruction-tuned models to specialized domains often relies on supervised fine-tuning (SFT) on curated instruction-response examples, which is expensive to collect at scale. Synthetic training examples generated by a teacher LLM from a domain corpus can reduce this cost, but existing pipelines can produce homogenized outputs and do not consistently capture cross-passage or cross-document dependencies. We introduce EmbGen, a synthetic data generation pipeline that decomposes a corpus into entity-description pairs, reassembles them using semantic structure inferred from embedding similarity, and then generates question-answer (QA) pairs via proximity, intra-cluster, and inter-cluster sampling with cluster-specialized system prompts. We evaluate EmbGen against EntiGraph, InstructLab and Knowledge-Instruct on three datasets of varied semantic heterogeneity, under fixed token budgets (5 and 20 million tokens). We use lexical overlap metrics, an LLM-as-a-judge rubric, and Binary Accuracy, a composed metric combining Factual Accuracy and Completeness for evaluation. EmbGen improves Binary Accuracy on the most heterogeneous dataset by 12.5\% at 5M and 88.9\% at 20M tokens budget, relative to the strongest baseline, while remaining competitive across other datasets with lower heterogeneity.
\end{abstract}


\maketitle

\section{Introduction}
Large language models are increasingly deployed under tight latency, memory, and cost constraints, making small instruction-tuned models attractive for many real applications \cite{hsieh_distilling_2023, wan_efficient_2024}. However, adapting a small model to a specialized domain often relies on SFT on curated instruction–response examples, which can be expensive to collect at scale \citep{riabi_synthetic_2021,mekala_leveraging_2022,ouyang_training_2022,wang_self-instruct_2023,huang_datagen_2025}. A common alternative is to generate synthetic training examples from a domain corpus using a teacher LLM \citep{ovadia_k-instruct_2025,ye_calibrating_2025}. In practice, without careful constraints and filtering, synthetic augmentation can yield homogenized outputs, factual inaccuracies, and insufficient coverage of long-tail domain content \cite{chen_graphgen_2025, shumailov_curse_2024, wang_self-instruct_2023}. Moreover, existing synthetic data pipelines often struggle to consistently capture cross-passage or cross-document (multi-hop) dependencies \citep{chen_graphgen_2025,ma_synthesize--graph_2025,wang_kcs_2025}, motivating generation pipelines that decompose the corpus into reusable units and reassemble them using semantic structure inferred from embedding similarity.

This paper introduces EmbGen, a data generation framework that converts a raw corpus into a structured synthetic SFT dataset by (i) decomposing documents into entity-description (ED) pairs, (ii) organizing ED pairs via embedding-based clustering and proximity grouping, and (iii) generating QA pairs through structure-aware sampling and cluster-specialized system prompts. Unlike approaches that rely on explicit curated knowledge graphs or generate QA pairs directly from contiguous chunks (\cite{chen_graphgen_2025,yang_scp_2024}), EmbGen uses lightweight adjacency graphs induced by embedding similarity to assemble semantically coherent generation contexts, even when relevant information is dispersed across different documents.

We frame EmbGen as a data-centric contribution by providing a reproducible pipeline and evaluation protocol for generating and evaluating synthetic SFT data from domain corpora under limited labelled supervision. Because lexical overlap metrics can underestimate gains in factual accuracy and completeness when responses are paraphrased or longer form, we report both lexical metrics and an LLM-based rubric emphasizing factual accuracy and completeness (\cite{liu_g-eval_2023,sellam_bleurt_2020,zhang_bertscore_2020}). Concretely, we evaluate EmbGen against three prior synthetic data methods: Knowledge-Instruct \citep{ovadia_k-instruct_2025}, InstructLab \citep{sudalairaj_lab_2024}, and EntiGraph (\cite{yang_scp_2024}), across corpora with varying heterogeneity. Results indicate that EmbGen can outperform other synthetic data generation methods as the heterogeneity of knowledge covered by a dataset increases. To assess these dimensions simultaneously, we report an LLM-judge Binary Accuracy that counts an answer correct only when Factual Accuracy is strong and Completeness is at least adequate. On Wikitext-10, the most heterogenous dataset in this study, EmbGen achieves relative Binary Accuracy gains of 12.5\% and 88.9\% under the LLM-as-a-judge rubric for 5M and 20M tokens budgets respectively.

Our work makes the following key contributions:
\begin{itemize}
    \item \textbf{EmbGen framework}. We propose an embedding-guided synthetic QA generation pipeline that constructs proximity groups and supports three sampling modes (proximity, intra-cluster and inter-cluster) together with cluster-specific system prompts.
    \item \textbf{Data-centric evaluation protocol}. We provide an evaluation methodology designed to test whether SFT data improves a model’s ability to answer questions requiring integration of evidence across multiple corpus documents, combining quantitative overlap metrics with an LLM-judge rubric focused on factual accuracy.
    \item \textbf{Benchmark study across heterogeneity regimes}. We present a controlled comparison across multiple datasets and token budgets, analyzing when structured semantic grouping and inter-cluster sampling help or hurt.
    \item \textbf{Reusable artifacts}. Upon manuscript acceptance, we will release (i) the EmbGen codebase, (ii) generated synthetic QA datasets, (iii) evaluation sets and rubric prompts, and (iv) scripts and configurations to reproduce baselines and results.
\end{itemize}

\section{Related Work}
Our work connects three active lines of research: domain adaptation via synthetic data generation, graph-guided data generation, and LLM-based data augmentation. We summarise each line of work and then position EmbGen.

\subsection{Domain Adaptation via Synthetic Data generation}
Recent methods show that a domain corpus can be amplified into a much larger synthetic training set for domain adaptation \citep{chen_graphgen_2025, ovadia_k-instruct_2025, sudalairaj_lab_2024, yang_scp_2024}. Entigraph constructs an entity-centric graph from a small corpus and samples entity paths, using an LLM to turn these paths into plain text for continued pretraining \cite{yang_scp_2024}. This synthetic corpus improves closed-book QA and instruction following on the target domain. Knowledge-Instruct generates information-dense synthetic instruction-response pairs and performs instruciton tuning for knowledge injection, as an alternative to raw text continued pretraining (CPT) for instruction-runed chat models \cite{ovadia_k-instruct_2025}. Similarly, InstructLab uses a taxonomy-guided synthetic data pipeline and multi-phase tuning to fine-tune LLMs, generating instruction-response data (including QA-style “knowledge” examples) from a curated taxonomy \cite{sudalairaj_lab_2024}. For the knowledge branch of the taxonomy, generation is grounded in source documents provided by the curator \cite{sudalairaj_lab_2024}. These methods generate synthetic corpora or instruction-response data for adaptation. EmbGen differs by inducing corpus structure from embeddings (via proximity groups) and generating QA pairs from those groups, rather than relying on curated taxonomies or knowledge graphs.

\subsection{Graph-Driven Data Generation and Representation}
Other work employs explicit structure to guide synthetic QA creation. Entigraph extracts entities to build a Knowledge Graph (KG) \cite{yang_scp_2024}. GraphGen constructs a KG from source documents, identifies knowledge gaps, and generates atomic and multi-hop QA pairs through graph sampling \cite{chen_graphgen_2025}. LEADING uses text-attributed graphs (TAGs), where each node has an associated text chunk and the edges define its neighborhood \cite{xue_efficient_2024}. It jointly fine-tunes a language model to encode the node text and a GNN to propagate those representations over the fixed graph during training \cite{xue_efficient_2024}. KG-SFT augments existing QA datasets by constructing reasoning subgraphs and generating natural-language explanations \cite{chen_kg_sft_2025}. In specialized domains, FinLLMs builds formula graphs to template numerical reasoning questions \cite{yuan_finllms_2024}. StrucSum builds a sentence-level similarity graph and uses simple graph cues (e.g., neighbors and centrality) to decide what to include in prompts for zero shot extractive summarization, without updating the underlying language model \cite{yuan_strucsum_2026}.

These methods highlight the utility of structured representations such as KGs, but many depend on curated graphs or existing QA datasets. EmbGen instead starts from a raw corpus and induces lightweight adjacency graphs from embeddings, so semantically related entities can drive QA generation without explicitly construct a KG.

\subsection{LLM-Based Data Augmentation}
Most synthetic data pipelines rely on global templates or small prompt sets for LLM generation \citep{yang_scp_2024,chen_kg_sft_2025,ovadia_k-instruct_2025}. This approach ignores the heterogeneity within a domain: historical narratives require different question styles and answer structures than technical definitions or procedural descriptions. EmbGen introduces cluster-specific prompting, providing the teacher model with local corpus context when generating QA pairs. This encourages questions and answers that align with the semantic neighbourhood from which the entities were drawn, with the intent of reducing generic or off-topic questions and encouraging answers that reuse the corpus’s domain-specific terminology and peculiarities.

\section{Methodology}
\begin{figure}[ht]
    \centering
    \includegraphics[width=\linewidth]{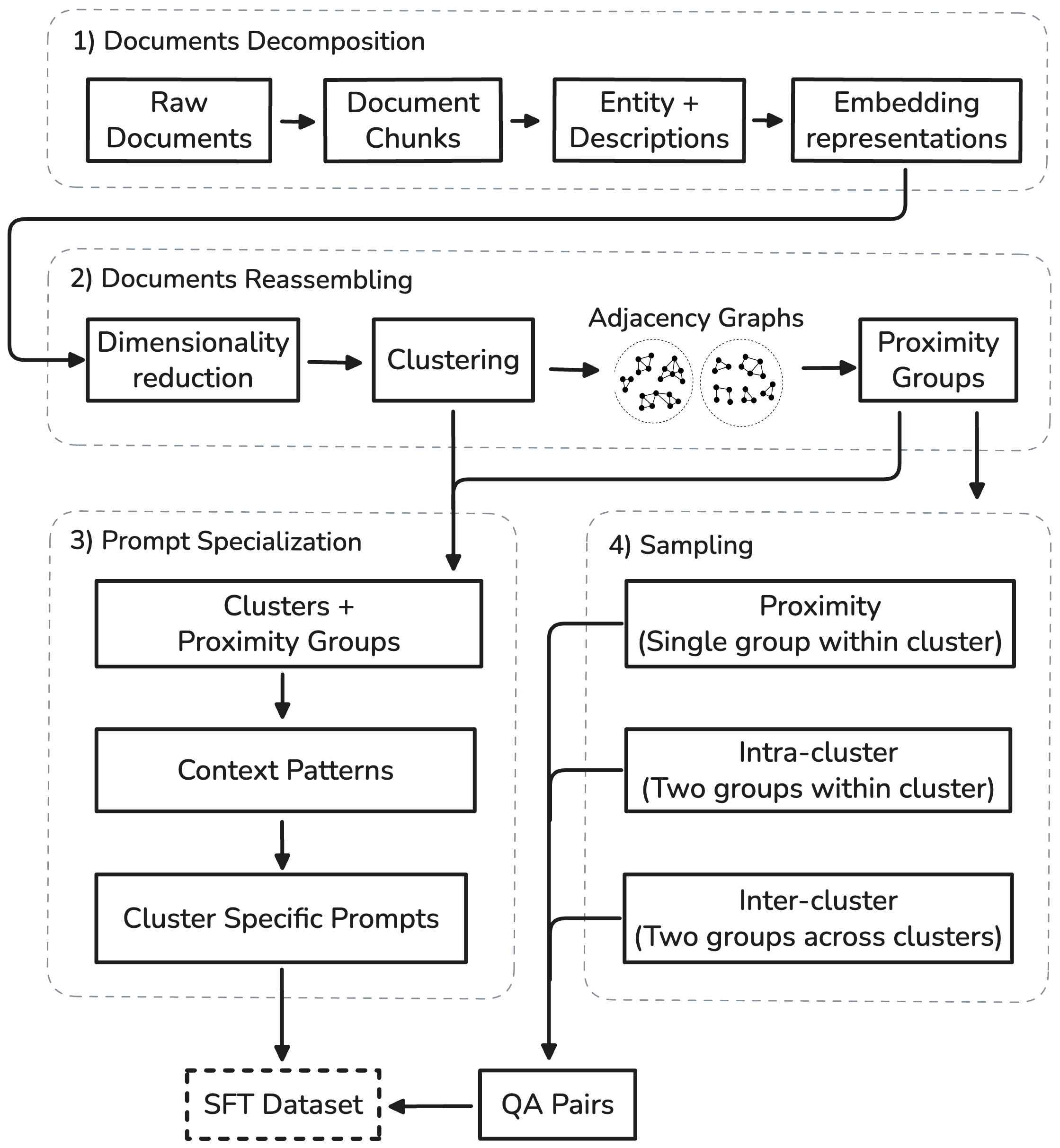}
    \caption{EmbGen decomposition-and-reassembling pipeline. (1) Raw documents are chunked, an LLM extracts ED pairs from each chunk and the concatenated entity and description text is encoded to produce ED pair embeddings; (2) embeddings undergo dimensionality reduction and clustering; adjacency graphs are then used to define proximity groups within clusters; (3) clusters and proximity groups are used to construct cluster-specific prompts; (4) proximity groups serve as inputs to three sampling strategies which are used to produce QA pairs.}
    \label{fig:embgen-pipeline}
\end{figure}

EmbGen generates synthetic QA pairs to improve the efficiency of SFT by restructuring a raw corpus $\mathcal{D}$ before generation. As shown in Figure~\ref{fig:embgen-pipeline}, Stages 1-2 implement a decomposition-and-reassembling mechanism: we first decompose documents into ED pairs and then reassemble these ED pairs into a semantic structure using embeddings, clustering, and proximity grouping. Stages 3-4 consume this reassembled structure to (i) construct cluster-specific system prompts and (ii) generate QA pairs from structured contexts (Figure~\ref{fig:embgen-pipeline}). Details of EmbGen hyperparameters are in Appendix~\ref{app:train_details} and prompt templates are specified in Appendix~\ref{app:prompts}.

\subsection{Stage 1: Decomposition into ED Pairs and Embeddings}
We segment the corpus $\mathcal{D}$ into chunks $\{x_t\}_{t=1}^{T}$ using a fixed chunking procedure applied uniformly across EmbGen and all baselines. For each chunk $x_t$, an LLM extracts ED pairs using a fixed entity extraction prompt (Appendix~\ref{app:prompts_entity_extraction}). Throughout EmbGen, we use a single teacher LLM ($M_{\text{teach}}$) across all datasets to execute all generation and curation prompts, including ED extraction and consolidation in Stage 1 and QA synthesis and prompt specialisation in later stages. Dataset-specific teacher model assignments are specified in Appendix~\ref{app:teacher_model_generation_settings}. Each extracted pair consists of an entity string and its corresponding contextualized textual description grounded strictly in $x_t$. All extracted pairs are pooled across the corpus, yielding a global, chunk-agnostic ED collection. This follows the same high-level entity-centric extraction paradigm used in EntiGraph \citep{yang_scp_2024}, with EmbGen additionally retaining explicit descriptions per entity.

Because extraction is independent across chunks, the pooled ED collection can contain duplicates, surface-form variants, and multiple descriptions for the same entity. We normalize entity strings using character TF-IDF similarity, and group highly similar names using a fixed threshold (Appendix~\ref{app:entity_norm_consolidation}). For each resulting group, we consolidate associated descriptions with a consolidation prompt, then apply a contradiction resolution prompt to select a single and consistent description (Appendix~\ref{app:prompts_consolidation}). This yields a one-to-one mapping between each canonical entity $e_i$ and its resolved description $d_i$, forming the consolidated set $\mathcal{E}=\{ED_i\}_{i=1}^{N}$ with $ED_i=(e_i,d_i)$. Implementation details for entity grouping and description resolution, including context bounds and fallback behaviour, are given in Appendix~\ref{app:entity_norm_consolidation}. After extraction and consolidation into $\mathcal{E}$, subsequent stages operate only on $\mathcal{E}$ and ignore the original chunking $\{x_t\}_{t=1}^{T}$ of $\mathcal{D}$.

For embedding, we concatenate $e_i$ and $d_i$ and encode the result with a sentence encoder to obtain $\mathbf{v}_i = f_{\mathrm{enc}}(\mathrm{concat}(e_i, d_i))$. We use all-mpnet-base-v2 \citep{song_mpnet_2020} consistently wherever embeddings are required in the pipeline (model configuration in Appendix~\ref{app:embedding_model_config}), although EmbGen is agnostic to the specific encoder choice.

\subsection{Stage 2: Reassembly into Proximity Groups}
This stage reassembles the embedded ED set $\mathcal{V} = \{\mathbf{v}_i\}_{i=1}^{N}$ into semantic contexts through clustering and constructing proximity groups within each cluster. 

\textbf{Dimensionality reduction and clustering.} We project each embedding to a lower dimension representation $\mathbf{z}_i = \mathrm{UMAP}(\mathbf{v}_i)$ to facilitate clustering. The reduced embeddings $\{\mathbf{z}_i\}$ are partitioned into $K$ clusters; we denote by $C_k \subseteq \{1,\dots,N\}$ the set of ED indices assigned to cluster $k$. Clustering either uses HDBSCAN \citep{campello_density-based_2013} or K-Means \citep{lloyd_least_1982}. For K-Means, the number of clusters $K$ is selected automatically via an inertia-based elbow criterion \citep{ketchen_jr_application_1996}. UMAP and Clustering parameters are reported in Appendix~\ref{app:dimensionality_reduction} and~\ref{app:clustering_config}.

\textbf{Proximity graph construction.} Within each cluster $C_k$, we construct an undirected proximity graph $G_k = (V_k, E_k)$ where $V_k = C_k$. An edge $(i, j) \in E_k$ exists if and only if $\cos(\mathbf{v}_i, \mathbf{v}_j) \geq \tau$, where $\tau$ is a similarity threshold. Edge construction uses the original high-dimensional embeddings $\mathbf{v}_i$ rather than the reduced representations $\mathbf{z}_i$, preserving fine-grained similarity information.

\textbf{Proximity group extraction.} Proximity groups $\{P_{k,m}\}$ are defined as the connected components of $G_k$. Two ED pairs belong to the same proximity group if and only if they are connected through a path of pairwise similarities exceeding $\tau$. Direct similarity between all pairs is not required. We enforce size constraints by splitting oversized components and expanding singletons through threshold adjustment (Appendix~\ref{app:proximity_group_construction}).

\subsection{Stage 3: System Prompt Specialisation}
In this stage, EmbGen constructs one cluster-specific system prompt $S_k$ per cluster $C_k$ using a shared base prompt $S_0$ (Appendix~\ref{app:Base_prompt}) and the proximity groups produced in Stage 2 (Figure~\ref{fig:embgen-pipeline}). This specialization adapts the system-level guidance so that it is more relevant to the subset of ED pairs represented by $C_k$, while preserving the task definition, output format, and behavioral constraints specified in $S_0$.

\textbf{Pattern extraction and consolidation.} For cluster $C_k$, EmbGen uses $M_{\text{teach}}$ to extract a bounded set of short textual descriptors (context patterns) from its proximity groups $\{P_{k,m}\}$ using a fixed extraction prompt (Appendix~\ref{app:Cluster_context_extraction_prompt}). Each newly extracted pattern is consolidated incrementally against the current set by $M_{\text{teach}}$ using a consolidation prompt (Appendix~\ref{app:Context_consolidation_prompt}) whereby a pattern is discarded as redundant, merged with an existing pattern, or added as a new entry. This continues until all $P_{k,m}$ have been processed or we reach a maximum of $L$ retained patterns for the cluster. The only Stage 3 hyperparameter is $L$, which we report in Appendix~\ref{app:Prompt_Specialisation}.

\textbf{Prompt specialization.} The consolidated patterns for $C_k$ are integrated into $S_0$ to produce $S_k$ by prompting $M_{\text{teach}}$ with the specialization prompt (Appendix~\ref{app:Cluster-specific_prompt_optimisation_prompt}). 

Stage 3 depends only on the proximity groups $\{P_{k,m}\}$ and cluster assignments and can execute independently of Stage 4 (Figure~\ref{fig:embgen-pipeline}).

\subsection{Stage 4: Context sampling and QA synthesis}
In Stage 4, EmbGen generates synthetic QA pairs by sampling proximity groups from Stage 2 and providing ED pairs contained in the selected group(s) to $M_{\text{teach}}$ (Figure~\ref{fig:embgen-pipeline}). We employ three sampling strategies described as follows.

\textbf{Proximity sampling.}  Each proximity group $P_{k,m}$ is passed to $M_{\text{teach}}$ exactly once as a standalone context, providing complete coverage of all reassembled units. Prompt templates differ based on whether $P_{k,m}$ contains one or multiple ED pairs (Appendix~\ref{app:prompts_proximity_QA_generation} and ~\ref{app:prompts_single_entity_fallback}). Proximity sampling executes first and yields an observed number of QA pairs $n_k^{\text{prox}}$ per cluster $C_k$.

\textbf{Intra-cluster sampling.} For each generation instance, we sample $g$ proximity groups from the same cluster (with $g=2$ in our experiments). The sampled groups $\{P_{k,m_1}, \ldots, P_{k,m_g}\}$ are passed to $M_{\text{teach}}$ using a multi-group prompt template that instructs the model to generate QA pairs leveraging relationships both within and across the provided groups (Appendix~\ref{app:prompts_multi_group}). Each cluster $C_k$ has a target QA count $n_k^{\text{intra}} = n_k^{\text{prox}} \cdot (\rho_{\text{intra}} / \rho_{\text{prox}})$. Sampling continues until the target is reached for each eligible cluster; clusters containing fewer than g proximity groups are excluded.

\textbf{Inter-cluster sampling.} We sample proximity groups from different clusters and combine them into a single generation context. When at least $g$ clusters exist ($K \geq g$), we sample $g$ distinct clusters with weights proportional to the number of proximity groups they contain, then draw one proximity group per selected cluster, yielding $\{P_{k_1,m_1}, \ldots, P_{k_g,m_g}\}$ where $k_i \neq k_j$. When fewer than $g$ clusters exist ($K < g$), clusters are sampled with replacement, allowing the same cluster to contribute multiple proximity groups; each selection still draws a new proximity group, preserving inter-cluster synthesis as long as at least two clusters exist. The sampled groups are passed to $M_{\text{teach}}$ using the multi-group prompt template (Appendix~\ref{app:prompts_multi_group}). The global target QA pair count is $n^{\text{inter}} = n^{\text{prox}} \cdot (\rho_{\text{inter}} / \rho_{\text{prox}})$, where $n^{\text{prox}} = \sum_k n_k^{\text{prox}}$.

Sampling effort is controlled by user-defined ratios $\rho_{\text{prox}} > 0$, $\rho_{\text{intra}} \geq 0$, and $\rho_{\text{inter}} \geq 0$ that sum to one. Rather than fixing the total QA pairs count in advance, EmbGen determines intra-cluster and inter-cluster targets after proximity sampling completes, using the observed counts as the scaling reference. Intra-cluster and inter-cluster sampling reuse the same proximity group pool without enumerating all possible combinations. All sampling parameters (including $g$ and $\rho_{\text{prox}}, \rho_{\text{intra}}, \rho_{\text{inter}}$) are reported in Appendix~\ref{app:synthetic_qa_generation}.

\subsection{Dataset Construction}
The final synthetic dataset is constructed by pairing each generated QA pair $(q, a)$ with a system prompt $S \in \{S_0\} \cup \{S_k\}_{k=1}^{K}$, chosen according to the proximity group(s) $\{P_{k,m}\}$ used as the generation context for $M_{\text{teach}}$ (Figure~\ref{fig:embgen-pipeline}). If all sampled groups belong to a single cluster $C_k$ (proximity sampling or intra-cluster sampling), we pair $(q, a)$ with $S_k$ to form $(S_k, q, a)$. If sampled groups spans multiple clusters (inter-cluster sampling), we pair $(q, a)$ with $S_0$ to form $(S_0, q, a)$, as no single cluster-specific prompt applies. The resulting dataset is a collection of $(S, q, a)$ triples. All prompts inherit the shared task definition from $S_0$.

\section{Experiments}
\subsection{Experimental Datasets}
To demonstrate the ability of EmbGen to effectively generate training data for SFT, we compared it to Knowledge-Instruct~\cite{ovadia_k-instruct_2025}, InstructLab~\cite{sudalairaj_lab_2024} and EntiGraph~\cite{yang_scp_2024}. These, along with EmbGen, were all evaluated on the following three datasets with varying degrees of heterogeneity, i.e. the level of semantic relatedness across documents. We empirically verified our heterogeneity assumptions by computing corpus-level similarity statistics (methods and results in Appendix~\ref{app:Dataset_Heterogeneity_Quantification}). Appendix~\ref{app:Clustering_Diagnostics} further reports clustering diagnostics (UMAP projections, cluster-size distributions, and inertia curves) to visualize the embedding-induced semantic neighborhoods used by EmbGen across the three corpora.

\subsubsection{Pop-QA-Cities-20.} A subset of Pop-QA and focuses on 20 European cities. Pop-QA-Cities-20 has the lowest level of heterogeneity relative to the other two datasets (Appendix~\ref{app:Dataset_Heterogeneity_Quantification}). The entity extraction phase generated 6,751 entities that the EmbGen augmentation used to produce 71M synthetic tokens from the 281k tokens in the 20 city entries taken from Pop-QA.

\subsubsection{SQuAD-20.} This dataset focuses on a specific subject matter and contains 20 documents chosen by Claude-Sonnet-4.5 from the Stanford Question Answer Dataset (SQuAD)~\cite{rajpurkar_squad_2016}. The goal was to recreate a scenario in which a corpus is made up of loosely semantically related documents. SQuAD-20 features documents regarding the United States, from institutional records (The Federal Bureau of Investigation) to contemporary popular culture (American Idol) and it has intermediate heterogeneity compared to the other two datasets (Appendix~\ref{app:Dataset_Heterogeneity_Quantification}). The EmbGen augmentation takes 4,452 extracted entities, produces 12M synthetic tokens from the 20 SQuAD articles which together contain 159k tokens.

\subsubsection{Wikitext-10.} A subset of the Salesforce Wikitext dataset that has been reduced to ten high quality articles by random sampling, intended to create a high-heterogeneity setting (Appendix~\ref{app:Dataset_Heterogeneity_Quantification}). The EmbGen augmentation extracts 2,107 entities to produce 5M synthetic tokens from the original Wikitext-10 corpus of about 2M tokens.

To assess generalisation between methods, we run all experiments under two token budget scenarios (5M and 20M). Using a fixed token budget ensures a fair comparison in terms of computational cost across methods.

\subsection{Training Conditions}
For our SFT conditions, we formatted groups of QA pairs and context rich prompts datasets into an instruction templated format to achieve consistent alignment and improved performance with the question answering process~\cite{jiang_mistral_2023}.
The Llama-3-8B-Instruct model was used to evaluate all synthetic generation methods within a fine-tuning regimen based on Low-Rank Adaptation (LoRA,~\cite{hu_lora_2021}). Fine-tuning was applied to the query and value projection matrices in the self-attention mechanism, as well as to the lower (down-projection) layer of the multi-layer perceptron block with an AdamW optimizer~\cite{loshchilov_decoupled_2019}.
We used a warmup ratio rather than a fixed number of warmup steps to account for the variation in the effective batch sizes during SFT, which is acceptable to maintain the QA alignment from the pretrained base model. The number of tokens per batch ranged from approximately 8k to 30k across experiments depending on the generation method, leading to differences in the total number of optimisation steps and training duration.
Details of training hyperparameters are shown in Appendix~\ref{app:Training_Parameter_Conditions}.

\subsection{Ablation Studies}
We conducted a focused set of ablation studies across all three datasets to assess the sensitivity of EmbGen to design choices that directly affect how semantic structure is constructed and exploited during synthetic QA generation. The intent of these experiments is diagnostic rather than exhaustive and is to isolate how changes to core components influence downstream behaviour rather than to identify a globally optimal configuration.
EmbGen has many configurable parameters across its stages, including clustering hyperparameters, similarity thresholds, proximity group size limits, expansion bounds, prompt construction settings, and sampling ratios. Exhaustive exploration of this space is computationally infeasible and is not the goal of this work. Instead, we selected parameters expected to have the largest qualitative impact on generated contexts and the diversity of resulting QA pairs.

Unless otherwise specified, all experiments use K-Means clustering, cluster-specific prompts $S_k$ with $L=5$ maximum patterns, and sampling ratios $\rho_{\text{prox}} = 0.6$, $\rho_{\text{intra}} = 0.3$, $\rho_{\text{inter}} = 0.1$. When a parameter is ablated, all others remain fixed.

\subsubsection{Ablated Parameters.} Table~\ref{tab:ablations} reports ablations over three parameter classes: prompt specialization, sampling strategy composition and clustering algorithm. Several EmbGen parameters are not ablated, including the maximum number of ED pairs per proximity group, similarity threshold $\tau$, group refinement bounds, and the number of proximity groups sampled per generation instance ($g$). These are fixed to values selected through preliminary experimentation, not claimed to be optimal, and define a stable operating regime for comparing the ablated settings.

\begin{table}[!hbp]
\centering
\caption{Ablation study settings.}
\label{tab:ablations}
\renewcommand{\arraystretch}{0.9} 
\small 
\begin{tabular}{lll}
\toprule
Ablation & Parameter Changed & Value \\
\midrule
Base prompt only
& Prompt type
& $S_0$ only (no $S_k$) \\

Reduced patterns
& Max patterns per $S_k$
& $L = 1$ (vs.\ 5) \\

No inter-cluster
& Sampling ratios
& \begin{tabular}[t]{@{}l@{}}
  $\rho_{\text{prox}} = 0.6,$\\
  $\rho_{\text{intra}} = 0.4,$\\
  $\rho_{\text{inter}} = 0.0$
  \end{tabular} \\

No intra-cluster
& Sampling ratios
& \begin{tabular}[t]{@{}l@{}}
  $\rho_{\text{prox}} = 0.6,$\\
  $\rho_{\text{intra}} = 0.0,$\\
  $\rho_{\text{inter}} = 0.4$
  \end{tabular} \\

Density clustering
& Clustering method
& HDBSCAN (vs.\ K-Means) \\
\bottomrule
\end{tabular}
\label{tab:ablation}
\end{table}

\section{Evaluation}
\subsection{Evaluation Data}
Because the goal of EmbGen is to improve the fine-tuned model’s ability to integrate knowledge dispersed across the corpus (e.g., different chunks or documents), our evaluation questions must require reasoning over multiple text segments. However, many publicly available evaluation sets for document-grounded question answering do not contain sufficiently challenging QA pairs for this purpose. Their questions are often short factoids or can be answered using a single localised piece of information, rather than requiring the model to combine evidence drawn from different parts of the corpus. We regard this model capability as a desirable one for practical applications using LLMs in production contexts. For this reason, we constructed evaluation sets using an LLM, making sure to use a model (Claude-Sonnet-4.5) distinct from the one used to generate the training QA pairs and prompts to reduce potential biases. The evaluation sets are built using the corpus to produce QA pairs that frequently require reasoning over information distributed across multiple chunks, pages, or documents. An evaluation dataset was not built for the SQuAD-20 dataset because there was already a sufficiently diverse evaluation set included with the Stanford release that covered multiple facts per generation unit \cite{rajpurkar_squad_2016}. All evaluation datasets consist of  250 QA pairs each.

\subsection{Evaluation Methodology}
We evaluate answers produced by fine-tuned models using a combination of automatic similarity metrics and an LLM-based answer quality judge. The two classes of metrics serve complementary roles. Automatic metrics provide a coarse measure of lexical and semantic overlap with reference answers, while the LLM-based judge is used to assess correctness and coverage. This combined evaluation protocol is designed to test whether improvements in semantic structure and sampling lead to more accurate and complete answers, rather than merely increasing lexical overlap.

\subsubsection{Quantitative Measures}
To assess lexical and semantic similarity between each model-generated answer and the corresponding ground-truth reference, we employ four established quantitative metrics: BLEU-n (n = 1, 2, 4) \cite{papineni_bleu_2002}, ROUGE-1, ROUGE-2, ROUGE-L~\cite{lin_rouge_2004}, and METEOR~\cite{banerjee_meteor_2005}. BLEU and METEOR are computed using NLTK implementations~\cite{loper_nltk_2002}, while ROUGE is computed using the \textit{rouge\_score} python library with stemming enabled. These metrics provide a baseline assessment of lexical overlap and surface-level semantic alignment. However, they are inherently limited in several respects. Answers that are factually correct but heavily paraphrased may receive low scores, while longer answers that improve coverage and completeness may be penalized relative to concise references. Moreover, high token overlap does not guarantee factual correctness.

\subsubsection{LLM-based Answer Quality Judge}
To address the limitations of automatic metrics, we employ an LLM-based answer quality judge that uses a structured rubric to evaluate each predicted answer given its question and reference answer, assigning scores along four dimensions (Appendix~\ref{app:Evaluation_Prompts}; \cite{zheng_judging_2023}). The judge is instructed to rely only on the provided inputs and not to use external knowledge or assumptions. Each dimension is scored on a three-level scale (Strong, Adequate, Weak), with calibration guidelines for multi-part questions, minor numerical imprecision, verbosity, etc. provided in Appendix~\ref{app:Evaluation_Prompts}. The rubric consists of the following dimensions:
\begin{enumerate}
  \item Factual Accuracy: Evaluates whether the factual claims explicitly stated in the predicted answer are consistent with the reference answer. This dimension assesses correctness of stated content only and does not penalize answers for missing information. Answers containing unsupported or contradictory claims are penalized accordingly.
  \item Completeness: Assesses whether the predicted answer addresses all required elements of the question, including multi-part components. Missing secondary elements may result in an Adequate score, while omission of essential elements results in a Weak score.
  \item Relevance: Measures whether the answer remains focused on the question and avoids unnecessary or distracting information that does not contribute to answering it.
  \item Clarity: Evaluates the coherence, organization, and readability of the answer, independent of factual correctness. 
\end{enumerate}

We use GPT-5 as the judge model, distinct from the model used to build the evaluation sets to avoid bias.

From the judge outputs, we derive a binary accuracy indicator reflecting strict correctness under the reference answer. An answer is counted as correct if and only if its Factual Accuracy score is Strong and its Completeness score is either Strong or Adequate. This definition enforces strict factual consistency while allowing minor omissions that do not substantially alter whether the question is answered. Binary Accuracy is computed deterministically from the judge’s categorical scores and is not produced directly by the judge.

To improve score stability, we evaluate each QA pair with 10 judge runs using the same judge model (temperature = 0) and report the average. Categorical scores are mapped to ordinal values, averaged across repetitions, and converted back to categorical labels via rounding. Binary Accuracy is aggregated by averaging per-run binary outcomes and rounding to the nearest integer.

For the sake of clarity, we selected the best performing EmbGen ablation for each dataset and token size combination according to Binary Accuracy. In practice, conducting similar ablation studies is standard procedure for determining the optimal hyperparameter configuration.

\section{Results}
\subsection{NLP Metrics}
On overlap metrics (BLEU, ROUGE, METEOR), EmbGen yields higher BLEU and ROUGE scores than the no-augmentation baseline in nearly all cases, though METEOR improvements are not consistent (Table~\ref{tab:nlp-metrics}). Knowledge-Instruct leads on BLEU-1 across most settings, while InstructLab is strongest on higher-order BLEU and ROUGE variants, particularly on SQuAD-20 where it achieves best scores on all metrics. EmbGen is typically second-best on ROUGE metrics: at 20M on Pop-QA-Cities-20, EmbGen achieves second-best BLEU-2 (0.122), ROUGE-1 (0.338), ROUGE-2 (0.144), and ROUGE-L (0.284). On Wikitext-10 at 20M, EmbGen reaches best on ROUGE-2 (0.126) and ROUGE-L (0.253), with second-best ROUGE-1 (0.335), BLEU-1 (0.240), BLEU-2 (0.139) and BLEU-4 (0.062). Scaling reduces some gaps: on Wikitext-10, the BLEU-1 difference between EmbGen and Knowledge-Instruct narrows from 0.068 at 5M (0.209 vs 0.277) to 0.030 at 20M (0.240 vs 0.270). Given the mismatch between lexical overlap and factual correctness, we emphasize the LLM-as-a-judge results next.

\begin{table*}[t]
\centering
\caption{NLP metrics performance across benchmark datasets. The best EmbGen ablations were selected based on Binary Accuracy for each dataset and data size combination (see Table~\ref{tab:ablations-binary}). Best score highlighted in grey, second best in bold.}
\label{tab:nlp-metrics}
\renewcommand{\arraystretch}{0.82} 
\small 
\begin{tabular}{llrrrrrrr}
\toprule
Dataset & Method & BLEU-1 & BLEU-2 & BLEU-4 & ROUGE-1 & ROUGE-2 & ROUGE-L & METEOR \\
\midrule
\multicolumn{9}{l}{\textbf{Size: 5M}} \\
\addlinespace[0.5ex]
Pop-QA-Cities-20 & Baseline: llama 8B Instruct & 0.108 & 0.063 & 0.026 & 0.182 & 0.068 & 0.146 & 0.251 \\
 & EntiGraph & 0.135 & 0.076 & 0.028 & 0.203 & 0.076 & 0.153 & 0.237 \\
 & InstructLab & \textbf{0.221} & \cellcolor{gray!20} 0.144 & \cellcolor{gray!20} 0.077 & \cellcolor{gray!20} 0.363 & \cellcolor{gray!20} 0.172 & \cellcolor{gray!20} 0.309 & \textbf{0.258} \\
 & Knowledge-Instruct & \cellcolor{gray!20} 0.242 & \textbf{0.141} & \textbf{0.064} & \textbf{0.320} & \textbf{0.133} & \textbf{0.257} & \cellcolor{gray!20} 0.307 \\
 & EmbGen & 0.165 & 0.093 & 0.044 & 0.292 & 0.106 & 0.242 & 0.218 \\
\addlinespace
SQuAD-20 & Baseline: llama 8B Instruct & 0.019 & 0.010 & 0.004 & 0.041 & 0.014 & 0.039 & 0.083 \\
 & EntiGraph Squad 5M & 0.024 & 0.013 & 0.006 & 0.052 & 0.019 & 0.049 & 0.093 \\
 & InstructLab & \cellcolor{gray!20} 0.051 & \cellcolor{gray!20} 0.028 & \cellcolor{gray!20} 0.013 & \cellcolor{gray!20} 0.131 & \cellcolor{gray!20} 0.051 & \cellcolor{gray!20} 0.126 & \cellcolor{gray!20} 0.133 \\
 & Knowledge-Instruct & 0.036 & 0.018 & 0.008 & 0.069 & 0.022 & 0.065 & \textbf{0.119} \\
 & EmbGen & \textbf{0.043} & \textbf{0.023} & \textbf{0.010} & \textbf{0.117} & \textbf{0.044} & \textbf{0.112} & 0.118 \\
\addlinespace
Wikitext-10 & Baseline: llama 8B Instruct & 0.169 & 0.092 & 0.035 & 0.242 & 0.079 & 0.170 & \cellcolor{gray!20} 0.269 \\
 & EntiGraph & 0.200 & 0.092 & 0.030 & 0.264 & 0.070 & 0.187 & 0.169 \\
 & InstructLab & \textbf{0.212} & \textbf{0.133} & \cellcolor{gray!20} 0.064 & \textbf{0.335} & \cellcolor{gray!20} 0.139 & \cellcolor{gray!20} 0.262 & 0.202 \\
 & Knowledge-Instruct & \cellcolor{gray!20} 0.277 & \cellcolor{gray!20} 0.155 & \textbf{0.063} & \cellcolor{gray!20} 0.342 & 0.119 & 0.243 & \textbf{0.249} \\
 & EmbGen & 0.209 & 0.128 & 0.060 & 0.327 & \textbf{0.132} & \textbf{0.254} & 0.201 \\
\addlinespace
\addlinespace
\multicolumn{9}{l}{\textbf{Size: 20M}} \\
\addlinespace[0.5ex]
Pop-QA-Cities-20 & Baseline: llama 8B Instruct & 0.108 & 0.063 & 0.026 & 0.182 & 0.068 & 0.146 & 0.251 \\
 & EntiGraph & 0.128 & 0.071 & 0.028 & 0.195 & 0.071 & 0.148 & 0.224 \\
 & InstructLab & \textbf{0.213} & \cellcolor{gray!20} 0.141 & \cellcolor{gray!20} 0.073 & \cellcolor{gray!20} 0.356 & \cellcolor{gray!20} 0.173 & \cellcolor{gray!20} 0.305 & \textbf{0.261} \\
 & Knowledge-Instruct & \cellcolor{gray!20} 0.239 & \cellcolor{gray!20} 0.141 & \textbf{0.065} & 0.310 & 0.129 & 0.246 & \cellcolor{gray!20} 0.308 \\
 & EmbGen & 0.201 & \textbf{0.122} & 0.059 & \textbf{0.338} & \textbf{0.144} & \textbf{0.284} & 0.228 \\
\addlinespace
SQuAD-20 & Baseline: llama 8B Instruct & 0.019 & 0.010 & 0.004 & 0.041 & 0.014 & 0.039 & 0.083 \\
 & EntiGraph & 0.017 & 0.009 & 0.004 & 0.038 & 0.011 & 0.036 & 0.074 \\
 & InstructLab & \cellcolor{gray!20} 0.050 & \cellcolor{gray!20} 0.026 & \cellcolor{gray!20} 0.011 & \cellcolor{gray!20} 0.132 & \cellcolor{gray!20} 0.050 & \cellcolor{gray!20} 0.129 & \cellcolor{gray!20} 0.130 \\
 & Knowledge-Instruct & 0.036 & 0.019 & 0.009 & 0.069 & 0.024 & 0.065 & \textbf{0.122} \\
 & EmbGen & \textbf{0.043} & \textbf{0.023} & \textbf{0.010} & \textbf{0.119} & \textbf{0.045} & \textbf{0.114} & 0.117 \\
\addlinespace
Wikitext-10 & Baseline: llama 8B Instruct & 0.169 & 0.092 & 0.035 & 0.242 & 0.079 & 0.170 & \cellcolor{gray!20} 0.269 \\
 & EntiGraph & 0.173 & 0.076 & 0.026 & 0.231 & 0.062 & 0.169 & 0.152 \\
 & InstructLab & 0.202 & 0.124 & 0.059 & 0.321 & \textbf{0.124} & \textbf{0.250} & 0.196 \\
 & Knowledge-Instruct & \cellcolor{gray!20} 0.270 & \cellcolor{gray!20} 0.153 & \cellcolor{gray!20} 0.064 & \cellcolor{gray!20} 0.340 & 0.120 & 0.247 & \textbf{0.247} \\
 & EmbGen & \textbf{0.240} & \textbf{0.139} & \textbf{0.062} & \textbf{0.335} & \cellcolor{gray!20} 0.126 & \cellcolor{gray!20} 0.253 & 0.223 \\
\addlinespace
\addlinespace
\bottomrule
\end{tabular}
\normalsize
\FloatBarrier
\end{table*}

\subsection{LLM-as-a-Judge Metrics Description}
On the lower heterogeneity datasets, differences between methods are substantial (Table~\ref{tab:llm-judge-metrics}). At 5M tokens, InstructLab achieves the highest Binary Accuracy on both Pop-QA-Cities-20 (0.266) and SQuAD-20 (0.280), with EmbGen matching 0.266 on Pop-QA-Cities-20 and reaching 0.232 on SQuAD-20. Increasing the budget to 20M shifts the ranking in EmbGen’s favour, but the uplift is modest: EmbGen becomes best on Binary Accuracy for Pop-QA-Cities-20 (0.282 vs 0.266; +6.0\% relative) and SQuAD-20 (0.288 vs 0.280; +2.9\% relative), alongside the highest factual accuracy on SQuAD-20 at 20M (1.66) and tied highest factual accuracy on Pop-QA-Cities-20 at 20M (1.73).

On Wikitext-10, the most heterogeneous corpus, Binary Accuracy values are lower for all methods, but EmbGen remains best at both budgets (0.072 at 5M; 0.068 at 20M). Relative to the strongest competitor, this corresponds to a 12.5\% uplift at 5M (0.072 vs 0.064, EntiGraph) and an 88.9\% uplift at 20M (0.068 vs 0.036, InstructLab). This gap is consistent with a trade-off visible in the judge dimensions: some baselines achieve higher average Factual Accuracy on Wikitext-10 (e.g., EntiGraph 1.46 at 20M) but simultaneously drop sharply in Completeness (1.08 vs EmbGen’s 1.40). This suggests those baselines often answer only part of the question. By contrast, EmbGen covers more of each question’s required elements, yielding higher Completeness scores and a greater proportion of answers counted as correct under the strict Binary Accuracy criteria (Table~\ref{tab:llm-judge-metrics}).

\begin{table*}[!htbp]
\centering
\caption{LLM-as-a-Judge metrics performance across benchmark datasets. The best EmbGen ablations were selected based on Binary Accuracy for each dataset and data size combination (see Table~\ref{tab:ablations-binary}). Best score highlighted in grey, second best in bold.}
\label{tab:llm-judge-metrics}
\renewcommand{\arraystretch}{0.82} 
\small 
\begin{tabular}{llrrrrr}
\toprule
Dataset & Method & Factual Accuracy & Completeness & Relevance & Clarity & Binary Accuracy \\
\midrule
\multicolumn{7}{l}{\textbf{Size: 5M}} \\
\addlinespace[0.5ex]
Pop-QA-Cities-20 & Baseline: llama 8B Instruct & 1.25 & \textbf{1.98} & 2.15 & 2.85 & 0.042 \\
 & EntiGraph & 1.39 & \cellcolor{gray!20} 2.00 & 2.22 & 2.88 & 0.008 \\
 & InstructLab & \cellcolor{gray!20} 1.75 & 1.93 & \cellcolor{gray!20} 2.76 & \cellcolor{gray!20} 3.00 & \cellcolor{gray!20} 0.266 \\
 & Knowledge-Instruct & 1.54 & 1.94 & \textbf{2.59} & \textbf{2.96} & \textbf{0.079} \\
 & EmbGen & \textbf{1.68} & 1.93 & 2.53 & 2.83 & \cellcolor{gray!20} 0.266 \\
\addlinespace
SQuAD-20 & Baseline: llama 8B Instruct & 1.28 & 1.99 & 2.30 & 2.89 & 0.008 \\
 & EntiGraph Squad 5M & 1.29 & 1.95 & 2.35 & 2.90 & 0.036 \\
 & InstructLab & \cellcolor{gray!20} 1.65 & \cellcolor{gray!20} 2.14 & \cellcolor{gray!20} 2.90 & \textbf{2.98} & \cellcolor{gray!20} 0.280 \\
 & Knowledge-Instruct & 1.39 & 1.97 & 2.56 & \textbf{2.98} & 0.036 \\
 & EmbGen & \textbf{1.60} & \textbf{2.06} & \textbf{2.88} & \cellcolor{gray!20} 3.00 & \textbf{0.232} \\
\addlinespace
Wikitext-10 & Baseline: llama 8B Instruct & 1.05 & 1.38 & 2.02 & 2.90 & 0.000 \\
 & EntiGraph & \cellcolor{gray!20} 1.34 & 1.32 & 2.32 & 2.84 & \textbf{0.064} \\
 & InstructLab & 1.18 & \cellcolor{gray!20} 1.44 & \cellcolor{gray!20} 2.86 & \textbf{2.98} & 0.036 \\
 & Knowledge-Instruct & 1.19 & \textbf{1.42} & 2.67 & 2.96 & 0.032 \\
 & EmbGen & \textbf{1.32} & 1.38 & \textbf{2.76} & \cellcolor{gray!20} 2.98 & \cellcolor{gray!20} 0.072 \\
\addlinespace
\addlinespace
\multicolumn{7}{l}{\textbf{Size: 20M}} \\
\addlinespace[0.5ex]
Pop-QA-Cities-20 & Baseline: llama 8B Instruct & 1.25 & \textbf{1.98} & 2.15 & 2.85 & 0.042 \\
 & EntiGraph & 1.35 & \cellcolor{gray!20} 2.01 & 2.14 & 2.81 & 0.008 \\
 & InstructLab & \cellcolor{gray!20} 1.73 & 1.91 & \cellcolor{gray!20} 2.76 & \cellcolor{gray!20} 2.99 & \textbf{0.266} \\
 & Knowledge-Instruct & \textbf{1.54} & 1.95 & 2.60 & \cellcolor{gray!20} 2.99 & 0.054 \\
 & EmbGen & \cellcolor{gray!20} 1.73 & 1.92 & \textbf{2.69} & \textbf{2.96} & \cellcolor{gray!20} 0.282 \\
\addlinespace
SQuAD-20 & Baseline: llama 8B Instruct & 1.28 & 1.99 & 2.30 & 2.89 & 0.008 \\
 & EntiGraph & 1.37 & 2.04 & 2.24 & 2.91 & 0.040 \\
 & InstructLab & \textbf{1.65} & \cellcolor{gray!20} 2.09 & \cellcolor{gray!20} 2.94 & \cellcolor{gray!20} 3.00 & \textbf{0.280} \\
 & Knowledge-Instruct & 1.40 & 2.02 & 2.68 & 2.98 & 0.064 \\
 & EmbGen & \cellcolor{gray!20} 1.66 & \textbf{2.06} & \textbf{2.84} & \textbf{2.99} & \cellcolor{gray!20} 0.288 \\
\addlinespace
Wikitext-10 & Baseline: llama 8B Instruct & 1.05 & 1.38 & 2.02 & 2.90 & 0.000 \\
 & EntiGraph & \cellcolor{gray!20} 1.46 & 1.08 & 1.30 & 2.03 & 0.020 \\
 & InstructLab & 1.16 & \cellcolor{gray!20} 1.42 & \cellcolor{gray!20} 2.72 & 2.95 & \textbf{0.036} \\
 & Knowledge-Instruct & 1.20 & 1.36 & 2.59 & \cellcolor{gray!20} 2.97 & 0.020 \\
 & EmbGen & \textbf{1.25} & \textbf{1.40} & \textbf{2.62} & \textbf{2.97} & \cellcolor{gray!20} 0.068 \\
\addlinespace
\addlinespace
\bottomrule
\end{tabular}
\normalsize
\end{table*}

\subsection{Embgen Ablation Studies}
We evaluate sensitivity to clustering, sampling (proximity, intra and inter-cluster), and prompt specialization using Binary Accuracy (Table~\ref{tab:ablations-binary}), with full metric breakdowns in Appendix~\ref{app:Ablation_Results}. The best configuration is corpus and budget-dependent, but three patterns recur.\@ (i) Pop-QA-Cities-20: lighter inter-cluster mixing and limited prompt specialization can help, especially at larger budget, suggesting cluster signals are useful but keeping too many patterns is not.\@ (ii) SQuAD-20: removing prompt specialization is consistently beneficial; removing intra-cluster sampling causes the largest degradation (e.g.\ from 0.204 to 0.148 at 5M; from 0.244 to 0.136 at 20M), indicating within-cluster composition is important.\@ (iii) Wikitext-10: optimal settings shift with token budget and results are consistent with higher topic diversity and greater risk of semantic drift under heavy mixing.

Overall, the ablations show that the best performing configurations often simplify either the sampling mix (e.g.\ removing inter-cluster sampling) or the prompt specialisation, and the optimal configuration is dataset-dependent.

\begin{table}[!htbp]
\centering
\caption{Ablations results for Binary Accuracy. For each dataset and token budget we report results for the default EmbGen configuration (see section 4.3) and the ablations. Complete ablation results for all metrics are reported in Appendix~\ref{app:Ablation_Results}}
\label{tab:ablations-binary}
\renewcommand{\arraystretch}{0.82} 
\small 
\setlength{\tabcolsep}{4pt} 
\begin{tabular}{lp{2.6cm}cc}
\toprule
Dataset & Configuration & \makecell{5M} & \makecell{20M} \\
\midrule
Pop-QA-Cities-20
& Default & 0.261 & 0.249 \\
& No Intra-cluster & 0.178 & 0.017 \\
& No Inter-cluster & \textbf{0.266} & 0.266 \\
& HDBSCAN & 0.241 & 0.012 \\
& Base Prompt & 0.245 & 0.274 \\
& 1 Max Pattern & 0.237 & \textbf{0.282} \\
\addlinespace
SQuAD-20
& Default & 0.204 & 0.244 \\
& No Intra-cluster & 0.148 & 0.136 \\
& No Inter-cluster & 0.212 & 0.256 \\
& HDBSCAN & 0.192 & 0.260 \\
& Base Prompt & \textbf{0.232} & \textbf{0.288} \\
& 1 Max Pattern & 0.168 & 0.232 \\
\addlinespace
Wikitext-10
& Default & \textbf{0.072} & 0.036 \\
& No Intra-cluster & 0.040 & \textbf{0.068} \\
& No Inter-cluster & 0.020 & 0.032 \\
& HDBSCAN & 0.044 & 0.028 \\
& Base Prompt & 0.028 & 0.064 \\
& 1 Max Pattern & 0.028 & 0.044 \\
\bottomrule
\end{tabular}
\normalsize
\end{table}

\section{Discussion and Limitations}
EmbGen’s strongest gains appear under the LLM-as-a-judge rubric when strict correctness is required. This is most evident on Wikitext-10, where EmbGen attains the highest Binary Accuracy at both token budgets, partly by providing more complete answers than the baselines. On Pop-QA-Cities-20 and SQuAD-20, Binary Accuracy varies substantially across baselines, but EmbGen remains in the top tier and the ranking changes only modestly with a larger budget.

The results and ablations support the design choice to build generation context from embedding-based clustering and proximity grouping. This produces compact context bundles that are often semantically coherent. These bundles support QA generation and can be summarized into cluster-specific prompts. The sampling policy then enforces a direct trade off: broader coverage versus mixing in off-topic content. The ablations show no single best recipe. More mixing is not consistently better. Several best configurations reduce intra or inter-cluster sampling, depending on the corpus. Prompt specialization also behaves as a control knob rather than a default. On Pop-QA-Cities-20 at 20M tokens, conditioning on a single retained pattern performs best, while stronger pattern reduction degrades Binary Accuracy on other datasets.

Relative to baselines, the main difference is the level at which context is constructed. InstructLab’s synthetic data pipeline prompts a teacher LLM with source documents provided by a curator as grounding context, alongside local taxonomy seed examples~\cite{sudalairaj_lab_2024}. Knowledge-Instruct and EntiGraph preserve document locality, operating on intermediate units derived from individual documents~\cite{ovadia_k-instruct_2025, yang_scp_2024}. EmbGen disgregards document locality after ED extraction and assembles contexts from neighborhoods in embedding space. This may explain why overlap metrics and strict correctness do not always agree. EmbGen can generate more previously unseen QA pairs by sampling diverse proximity-based combinations of entities, yielding novel cross-entity interactions. In contrast, Knowledge-Instruct remains anchored to individual documents, rephrasing existing facts and converting them into instruction-style examples without synthesizing data across document boundaries.

This study has several limitations. EmbGen depends on an LLM for ED extraction and grounding; errors can propagate to grouping, sampling, and prompt specialization, and we do not quantify sensitivity. We ablate clustering choices and keep dimensionality reduction settings fixed for methodological consistency, but we do not study the sensitivity of EmbGen to the choice of embedding encoder (all-mpnet-base-v2~\cite{song_mpnet_2020} throughout). Robustness across alternative encoders remains for future work. EntiGraph is evaluated via a simplified SFT adaptation rather than its intended continued pretraining regime. Both data generation and scoring rely on LLMs, consequently judging can be biased (e.g. position and verbosity effects~\cite{zheng_judging_2023}), so human or multi judge checks would strengthen validity. Moreover, results are limited to the studied corpora, token budgets, and a single student model.

\section{Conclusions}
EmbGen is an embedding-guided pipeline for converting a raw corpus into a structured synthetic QA dataset. It decomposes documents into ED pairs, clusters semantically related ED pairs into proximity groups using embedding similarity, and then generates QA pairs from these groups, optionally using cluster-specific prompts to specialize the generation. In comparison to strong baseline methods, the results show that EmbGen is broadly competitive on overlap metrics, and its clearest benefits emerge when evaluated on strict correctness. As the corpus heterogeneity increases, EmbGen's grouping strategy yields notably more factually accurate and complete answers, leading to higher Binary Accuracy (for example, achieving up to nearly double the Binary Accuracy of the best baseline on the most heterogeneous dataset at the larger token budget).

The ablation study shows that the best hyperparameter settings vary by dataset, with no single configuration performing optimally across all cases. Often, the highest gains came from simplifying the generation strategy (e.g. reducing inter-cluster context) to suit a given corpus, rather than always using the maximum available structure.

Overall, the results position embedding-based grouping as a practical mechanism for structuring synthetic data generation. It enables the creation of compact, semantically coherent training examples that help models learn complex, cross-document knowledge with improved accuracy. Although our experiments focused on QA pair generation for fine-tuning, the underlying approach is quite general. In the future, EmbGen's proximity grouping paradigm could be applied beyond QA generation. For instance, to retrieval-augmented generation or other tasks that require assembling relevant context from a corpus to systematically construct better input contexts for language models.

\bibliographystyle{ACM-Reference-Format}
\bibliography{bib/complete-ref}

\clearpage
\appendix
\clearpage

\section{Parameterization of EmbGen}
\label{app:train_details}
This section documents the complete parameterization of EmbGen, covering entity consolidation, embedding generation, dimensionality reduction, clustering, proximity grouping, and synthetic question-answer generation. All parameters reported here correspond exactly to those used in the implementation. Unless explicitly stated otherwise, parameters are held fixed across all methods and ablation settings for a given dataset to ensure internal consistency and controlled comparisons.

\subsection{Entity Normalization and Consolidation}
\label{app:entity_norm_consolidation}
EmbGen performs similarity-based consolidation during entity validation to merge surface-level duplicates and near-duplicates produced by the extraction stage. Entity strings are first normalized by lowercasing and trimming whitespace. Similarity is computed using cosine similarity over character-level TF-IDF representations of normalized entity strings (character $n$-grams with $n \in \{1,2,3\}$). A similarity threshold of 0.85 determines whether two entities are grouped; merge decisions use transitive closure via Union-Find so that chains of above-threshold matches form a single group. Each group yields one canonical entity name, selected as the most frequent original surface form among its occurrences.

For each canonical entity, we aggregate all extracted descriptions and deduplicate them. If only one unique description exists, it is retained unchanged. If multiple distinct descriptions exist, we invoke $M_{\text{teach}}$: a consolidation prompt (Appendix~\ref{app:Entity_Consolidation_Prompt}) for similarity-merged groups, or a contradiction-resolution prompt (Appendix~\ref{app:Contradiction-Resolution_Prompt}) for single entities with conflicting descriptions. In both cases, we cap the number of unique candidate descriptions included in the LLM context at 10 to bound prompt length and reduce information dilution when consolidating many near-redundant variants. We did not tune this value; we selected 10 as a conservative fixed bound that balances coverage of description variants with prompt length constraints. Entities with more than 10 unique extracted descriptions are uncommon in our datasets, so this cap primarily serves as a practical safeguard for runtime and prompt budget rather than a tuned hyperparameter. If an LLM call fails or returns an empty output, we fall back to selecting the longest available description.

The similarity threshold and context limit are implementation choices, not derived from theoretical analysis or dataset-specific tuning. They are fixed across all datasets, methods, and ablations.

\subsection{Embedding Model Configuration}
\label{app:embedding_model_config}
Each entity is represented using a single embedding derived from a combined textual input consisting of the entity name and its explanation. Embeddings are generated using a transformer-based sentence embedding model with the following fixed configuration:
\begin{itemize}[label=\textbullet]
	\item Embedding model: all-mpnet-base-v2
	\item Batch size: 32
	\item Maximum token length: 512
	\item Automatic mixed precision: enabled
	\item Embedding normalization: enabled
	\item Pinned memory: enabled
\end{itemize}

\noindent
Embeddings are generated once per dataset and reused across all downstream stages, including dimensionality reduction, clustering, proximity grouping, and synthetic data generation.

\subsection{Dimensionality Reduction}
\label{app:dimensionality_reduction}
Prior to clustering, entity embeddings are projected into a lower-dimensional space using UMAP. This step is applied uniformly across all experiments and serves as an intermediate representation for subsequent clustering.

UMAP is configured with the following fixed parameters:
\begin{itemize}[label=\textbullet]
    \item Number of neighbours: 50
    \item Number of components: 15
    \item Minimum distance: 0.0
    \item Distance metric: cosine
    \item Random seed: 42
\end{itemize}

\noindent
These values are taken directly from the implementation and are not tuned per dataset or per method. They remain constant across all experiments to ensure that downstream behaviour is not confounded by changes in the dimensionality reduction stage.

\subsection{Clustering Configuration}
\label{app:clustering_config}
EmbGen supports both density-based and centroid-based clustering as primary configurations. Clustering is applied to the UMAP-reduced embeddings and instantiated using either HDBSCAN or K-Means, depending on the experimental setup.

For density-based clustering, HDBSCAN is configured with the following fixed parameters:
\begin{itemize}[label=\textbullet]
    \item Minimum cluster size: 100
    \item Minimum samples: 30
    \item Distance metric: Euclidean
    \item Cluster selection epsilon: 0.1
\end{itemize}

\noindent
These parameters control cluster density requirements and cluster merging behaviour. Noise points produced by HDBSCAN are treated as noise and excluded from downstream processes.

For centroid-based clustering, K-Means is used with automatic selection of the number of clusters:
\begin{itemize}[label=\textbullet]
    \item Cluster count range: [2, 100]
    \item Cluster selection method: elbow criterion
    \item Maximum number of candidate values evaluated: 50
    \item Initialization method: k-means++
    \item Maximum iterations: 300
    \item Random seed: 42
\end{itemize}

\noindent
Both HDBSCAN and K-Means are treated as first-class clustering methods within EmbGen and are evaluated under the same downstream pipeline components.

\subsection{Proximity Group Construction}
\label{app:proximity_group_construction}
Within each cluster, EmbGen constructs proximity groups based on cosine similarity between entity embeddings. Group construction begins from an initial similarity threshold and proceeds through controlled splitting and expansion.

The following parameters are fixed in the implementation:
\begin{itemize}[label=\textbullet]
    \item Initial similarity threshold $\tau$: 0.75
    \item Expansion step size: 0.01
    \item Maximum group size: 10
    \item Maximum expansion attempts for singleton entities: 10
    \item Minimum additional entities required to stop expansion: 1
\end{itemize}

\noindent
Groups exceeding the maximum size are split by incrementally increasing the similarity threshold. Singleton entities are expanded by progressively lowering the threshold until either additional neighbours are found or a lower bound is reached.

The lower bound for singleton expansion is 0.5, below which expansion is terminated. This bound is enforced uniformly across all experiments.

All proximity grouping parameters are held fixed unless explicitly ablated.

\subsection{Synthetic QA Generation}
\label{app:synthetic_qa_generation}
EmbGen generates synthetic QA pairs using three complementary strategies: proximity, intra-cluster, and inter-cluster generation. These strategies differ in how entity groups are sampled but share the same underlying grouping and clustering structure.

The relative contribution of each strategy is controlled using composition ratios:
\begin{itemize}[label=\textbullet]
    \item Proximity ratio $\in [0.0, 1.0]$
    \item Intra-cluster ratio $\in [0.0, 1.0]$
    \item Inter-cluster ratio $\in [0.0, 1.0]$
    \item Groups per generation instance ($g$): 2, fixed across all experiments.
\end{itemize}

\noindent
Subject to the constraint that the three ratios sum to one. Default ratios (unless otherwise specified): $\rho_{\text{prox}} = 0.6$, $\rho_{\text{intra}} = 0.3$, $\rho_{\text{inter}} = 0.1$.

All other parameters governing proximity group construction and QA generation are fixed as described above.

\subsection{Prompt Specialisation}
\label{app:Prompt_Specialisation}
In our default configuration, we set $L = 5$. In the ablation, we set $L = 1$ to vary maximum patterns extracted.

\subsection{Teacher Model and Generation Settings}
\label{app:teacher_model_generation_settings}
Synthetic QA generation in EmbGen is driven by a fixed teacher language model $M_{\text{teach}}$, chosen per dataset based on availability at the time each experimental batch was run:
\begin{itemize}[label=\textbullet]
    \item WikiText-10: $M_{\text{teach}}$ = gpt-4o-mini
    \item SQuAD-20 and Pop-QA-Cities-20: $M_{\text{teach}}$ = GPT-5
\end{itemize}

\noindent
This split reflects the chronological progression of experiments rather than an experimental variable. For a given dataset, $M_{\text{teach}}$ is held constant across all methods and ablations, so results are not confounded by changes in the generation model.

To further reduce variability, we fix decoding to a near-deterministic regime across all datasets and experiments:
\begin{itemize}[label=\textbullet]
    \item Temperature: 0.001
\end{itemize}

\noindent
This setting minimizes stochastic variation in outputs and is applied uniformly without dataset-specific tuning.

\subsection{Training Parameter Conditions}
\label{app:Training_Parameter_Conditions}
The number of training steps varies across datasets, as we calibrate the training regimen to fixed token budgets of 5M and 20M tokens per experiment. We do not employ sequence packing during fine-tuning in order to preserve the instruction-following alignment of the base model.

To determine the required number of training steps, we tokenize each generated QA dataset and compute the number of samples needed to reach the target token count. The effective batch size is computed as $B_{\text{eff}} = B_{\text{device}} \times N_{\text{devices}}$, where $B_{\text{device}} = 4$ and $N_{\text{devices}} = 8$, yielding $B_{\text{eff}} = 32$. The number of gradient updates is then obtained by dividing the sample count by $B_{\text{eff}}$. In cases where the target token budget is not exactly divisible by the batch size, we accept a tolerance of $\pm$20k tokens.

Table~\ref{tab:hyperparams} summarizes the fixed training hyperparameters used across all experiments.

\begin{table}[!htbp]
\centering
\caption{Training hyperparameters}
\small
\begin{tabular}{lc}
\toprule
Hyperparameter & Value \\
\midrule
$r$ & 16 \\
$\alpha$ & 64 \\
Dropout & 0.05 \\
Target layers & \texttt{[q\_proj, v\_proj, down\_proj]} \\
Learning rate & $2\times 10^{-5}$ \\
Learning rate schedule & Cosine \\
Warmup step ratio & 0.03 \\
A100 GPUs & 8 \\
Per-device batch size & 4 \\
\bottomrule
\end{tabular}
\label{tab:hyperparams}
\end{table}

\clearpage
\section{Experiments and Diagnostics}
\label{app:diagnostics}
\subsection{Dataset Heterogeneity Quantification}
\label{app:Dataset_Heterogeneity_Quantification}
We quantify dataset heterogeneity using the distribution of pairwise cosine similarities between entity-description embeddings within each corpus. For each dataset, we compute similarities over all unique entity pairs and summarize the resulting distribution using four statistics: mean, median, standard deviation, and interquartile range (IQR). Lower mean and median indicate lower average semantic relatedness across entities, which we interpret as higher corpus heterogeneity.

These statistics provide an empirical basis for treating Wikitext-10 as the most heterogeneous corpus among the three. Wikitext-10 exhibits the lowest mean and median similarity, along with the smallest interquartile range, indicating consistently low semantic relatedness across entities (Table~\ref{tab:summary-stats}). In contrast, Pop-QA-Cities-20 has the highest mean and median similarities, suggesting a more semantically concentrated corpus (Table~\ref{tab:summary-stats}). SQuAD-20 falls between these two extremes, consistent with our qualitative assessments in the main text and clustering diagnostics above (Table~\ref{tab:summary-stats}).

\begin{table}[htb]
\centering
\caption{Summary statistics of metric distributions across datasets. We report the mean, median, standard deviation (Std), and interquartile range (IQR) for each dataset.}
\label{tab:summary-stats}
\renewcommand{\arraystretch}{0.85}
\small
\begin{tabular}{lrrrr}
\toprule
Dataset & Mean & Median & Std & IQR \\
\midrule
Pop-QA-Cities-20 & 0.148724 & 0.135841 & 0.097579 & 0.114237 \\
SQuAD-20         & 0.121788 & 0.112446 & 0.082172 & 0.098247 \\
Wikitext-10      & 0.096303 & 0.087080 & 0.082128 & 0.091085 \\
\bottomrule
\end{tabular}
\normalsize
\end{table}

\subsection{Clustering Diagnostics}
\label{app:Clustering_Diagnostics}
\begin{figure}[!h]
\centering
\includegraphics[width=0.9\linewidth]{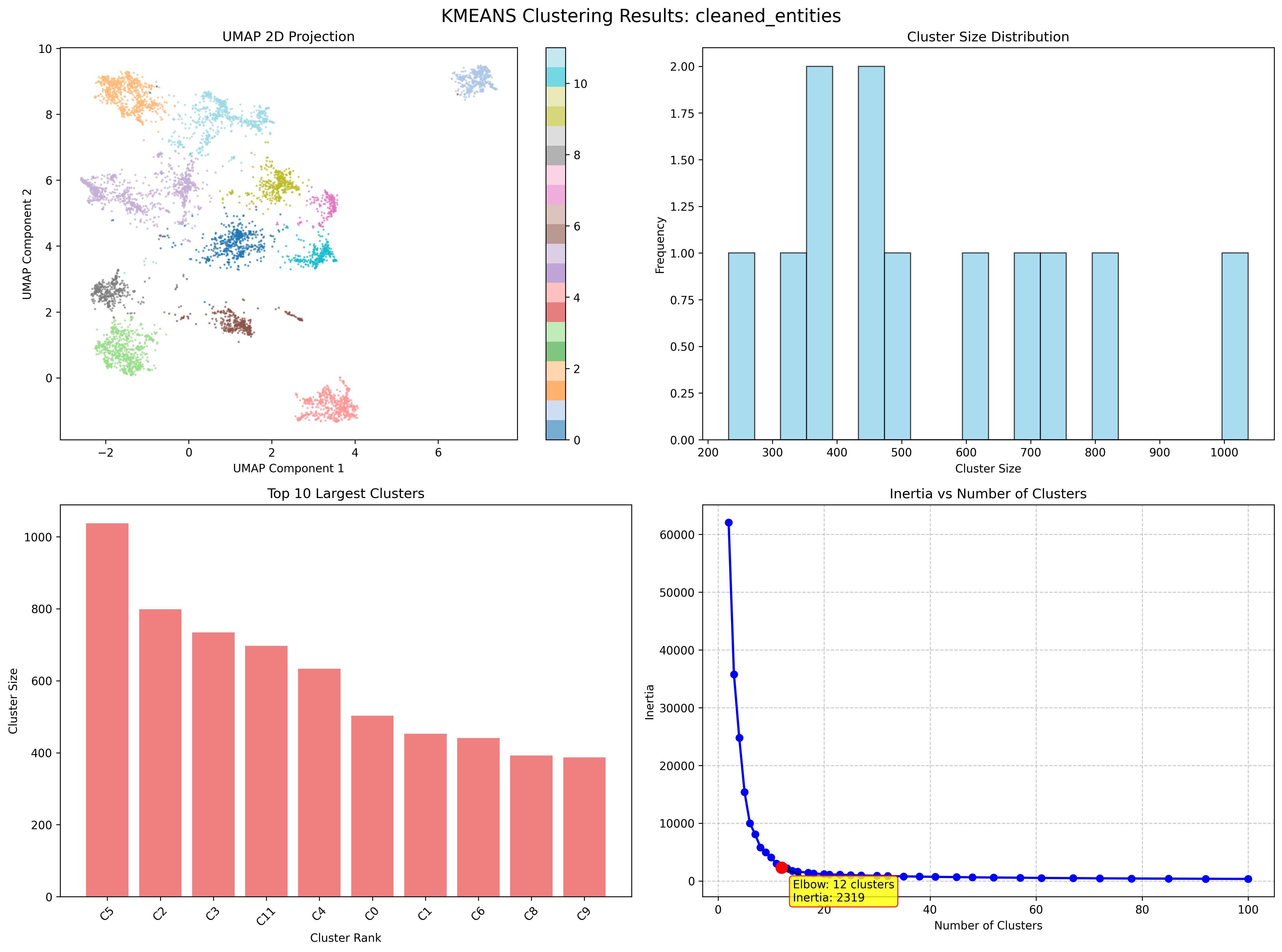}
\caption{Clustering and distribution diagnostics for POPQA-Cities}
\label{fig:popqa-cluster}
\end{figure}

\begin{figure}[!h]
\centering
\includegraphics[width=0.9\linewidth]{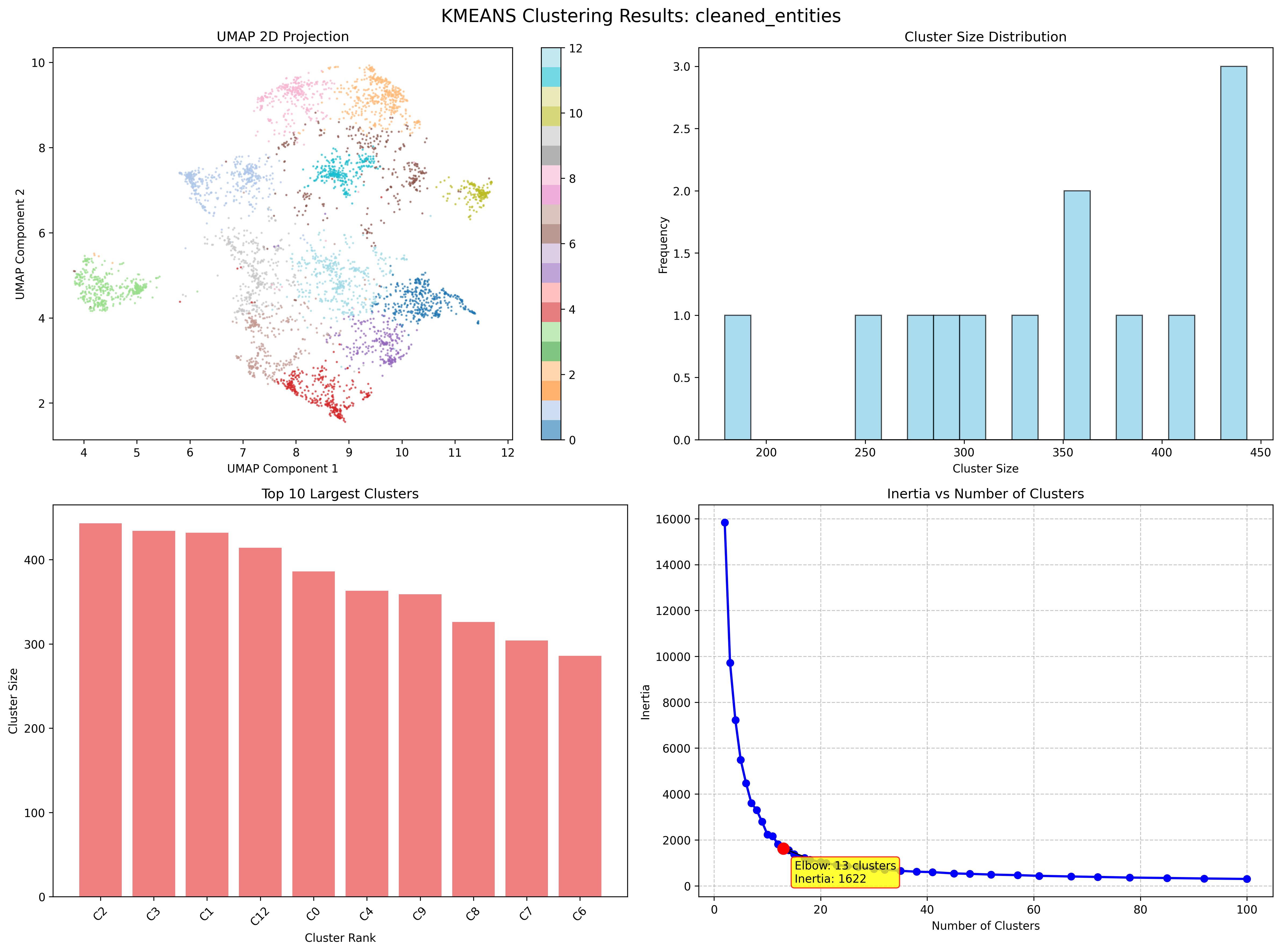}
\caption{Clustering and distribution diagnostics for SQuAD-20}
\label{fig:squad-cluster}
\end{figure}

\begin{figure}[!h]
\centering
\includegraphics[width=0.9\linewidth]{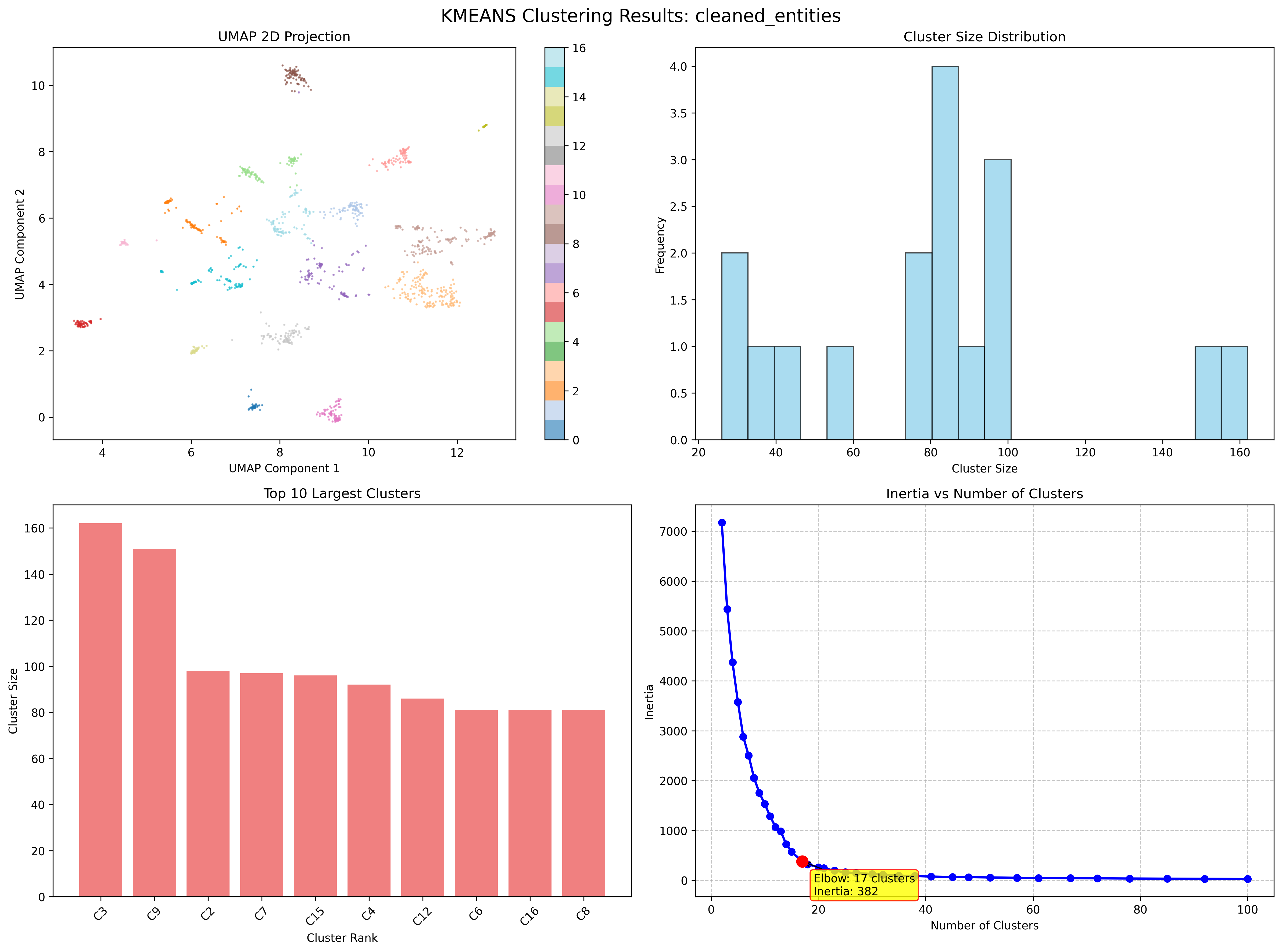}
\caption{Clustering and distribution diagnostics for Wikitext-10}
\label{fig:wikitext-cluster}
\end{figure}

This appendix provides qualitative diagnostics of the semantic structure EmbGen induces when reassembling corpora. For each dataset, we visualize (i) a 2D UMAP projection colored by cluster assignment, (ii) a histogram of cluster sizes, (iii) the sizes of the ten largest clusters, and (iv) the K-Means inertia curve over candidate cluster counts, with the elbow-selected value highlighted (Figures~\ref{fig:popqa-cluster},~\ref{fig:squad-cluster} and~\ref{fig:wikitext-cluster}). These plots are intended as diagnostics, not evaluation metrics. K-Means selects the number of clusters automatically using an inertia-based elbow criterion over a fixed candidate range, while density clustering variants (HDBSCAN) are configured with fixed parameters. UMAP is a qualitative projection that can distort global geometry, and cluster-size profiles depend on the chosen clustering criterion, but together they provide context for interpreting ablations that alter clustering and how clusters are exploited during generation.

\subsection {Ablation Results}
\label{app:Ablation_Results}
Tables~\ref{tab:nlp-metrics-full} and \ref{tab:llm-judge-metrics-full} report the full ablation results for all EmbGen variants across datasets and token budgets. Table~\ref{tab:nlp-metrics-full} provides the complete set of automatic overlap metrics (BLEU-1/2/4, ROUGE-1/2/L and METEOR). Table \ref{tab:llm-judge-metrics-full} reports the full LLM-as-a-judge rubric (Factual Accuracy, Completeness, Relevance and Clarity) and the derived Binary Accuracy.

\begin{table*}[!h]
\centering
\caption{Full quantitative overlap metrics. Full results for all EmbGen variants across datasets and token budgets using automatic overlap metrics.}
\label{tab:nlp-metrics-full}
\renewcommand{\arraystretch}{0.8} 
\small 
\begin{tabular}{llrrrrrrr}
\toprule
Dataset & Configuration & BLEU-1 & BLEU-2 & BLEU-4 & ROUGE-1 & ROUGE-2 & ROUGE-L & METEOR \\
\midrule
\multicolumn{9}{l}{\textbf{Size: 5M}} \\
\addlinespace[0.5ex]
Pop-QA-Cities-20 & Primary K-means             & 0.166 & 0.090 & 0.042 & 0.288 & 0.093 & 0.238 & 0.209 \\
                 & Ablation No Intra-cluster   & 0.172 & 0.098 & 0.043 & 0.287 & 0.108 & 0.231 & 0.245 \\
                 & Ablation No Inter-cluster   & 0.165 & 0.093 & 0.044 & 0.292 & 0.106 & \cellcolor{gray!20} 0.242 & 0.218 \\
                 & Ablation HDBSCAN            & \cellcolor{gray!20} 0.240 & \cellcolor{gray!20} 0.150 & \cellcolor{gray!20} 0.071 & \cellcolor{gray!20} 0.334 & \cellcolor{gray!20} 0.139 & 0.281 & \cellcolor{gray!20} 0.298 \\
                 & Ablation Base Prompt        & 0.211 & 0.125 & 0.060 & 0.337 & \cellcolor{gray!20} 0.142 & 0.281 & 0.243 \\
                 & Ablation 1 Max Pattern      & 0.189 & 0.108 & 0.047 & 0.308 & 0.114 & 0.246 & 0.232 \\
\addlinespace
SQuAD-20         & Primary K-means             & \cellcolor{gray!20} 0.048 & \cellcolor{gray!20} 0.025 & \cellcolor{gray!20} 0.012 & \cellcolor{gray!20} 0.125 & \cellcolor{gray!20} 0.047 & \cellcolor{gray!20} 0.122 & 0.117 \\
                 & Ablation No Intra-cluster   & 0.041 & 0.022 & 0.010 & 0.104 & 0.038 & 0.099 & 0.109 \\
                 & Ablation No Inter-cluster   & 0.044 & 0.024 & 0.011 & 0.117 & 0.045 & 0.114 & 0.117 \\
                 & Ablation HDBSCAN            & 0.040 & 0.022 & 0.010 & 0.105 & 0.040 & 0.101 & 0.112 \\
                 & Ablation Base Prompt        & 0.043 & 0.023 & 0.010 & 0.117 & 0.044 & 0.112 & \cellcolor{gray!20} 0.118 \\
                 & Ablation 1 Max Pattern      & 0.045 & 0.024 & 0.011 & 0.116 & 0.045 & 0.112 & 0.114 \\
\addlinespace
Wikitext-10      & Primary EmbGen K-means      & 0.209 & 0.128 & 0.060 & 0.327 & \cellcolor{gray!20} 0.132 & \cellcolor{gray!20} 0.254 & 0.201 \\
                 & Ablation No Intra-cluster   & 0.257 & 0.148 & 0.063 & \cellcolor{gray!20} 0.337 & 0.126 & 0.249 & 0.237 \\
                 & Ablation No Inter-cluster   & 0.267 & 0.146 & 0.060 & 0.324 & 0.113 & 0.239 & 0.247 \\
                 & Ablation HDBSCAN            & \cellcolor{gray!20} 0.276 & \cellcolor{gray!20} 0.151 & \cellcolor{gray!20} 0.064 & 0.331 & 0.119 & 0.243 & \cellcolor{gray!20} 0.251 \\
                 & Ablation Base Prompt        & 0.252 & 0.146 & 0.065 & 0.335 & 0.125 & 0.251 & 0.230 \\
                 & Ablation 1 Max Pattern      & 0.254 & 0.148 & \cellcolor{gray!20} 0.066 & 0.337 & 0.128 & 0.252 & 0.233 \\
\addlinespace[0.75ex]
\multicolumn{9}{l}{\textbf{Size: 20M}} \\
\addlinespace[0.5ex]
Pop-QA-Cities-20 & Primary K-means             & \cellcolor{gray!20} 0.203 & 0.120 & 0.056 & 0.337 & 0.133 & 0.281 & 0.238 \\
                 & Ablation No Intra-cluster   & 0.144 & 0.081 & 0.032 & 0.215 & 0.081 & 0.168 & \cellcolor{gray!20} 0.245 \\
                 & Ablation No Inter-cluster   & 0.189 & 0.111 & 0.051 & 0.318 & 0.124 & 0.261 & 0.228 \\
                 & Ablation HDBSCAN            & 0.084 & 0.046 & 0.016 & 0.121 & 0.041 & 0.092 & 0.155 \\
                 & Ablation Base Prompt        & 0.190 & 0.114 & 0.053 & 0.327 & 0.135 & 0.272 & 0.219 \\
                 & Ablation 1 Max Pattern      & 0.201 & \cellcolor{gray!20} 0.122 & \cellcolor{gray!20} 0.059 & \cellcolor{gray!20} 0.338 & \cellcolor{gray!20} 0.144 & \cellcolor{gray!20} 0.284 & 0.228 \\
\addlinespace
SQuAD-20         & Primary K-means             & 0.043 & \cellcolor{gray!20} 0.024 & 0.010 & 0.116 & \cellcolor{gray!20} 0.047 & 0.111 & 0.116 \\
                 & Ablation No Intra-cluster   & 0.038 & 0.021 & 0.009 & 0.099 & 0.038 & 0.095 & 0.115 \\
                 & Ablation No Inter-cluster   & \cellcolor{gray!20} 0.045 & 0.024 & \cellcolor{gray!20} 0.011 & \cellcolor{gray!20} 0.121 & 0.048 & \cellcolor{gray!20} 0.118 & 0.116 \\
                 & Ablation HDBSCAN            & 0.042 & 0.022 & 0.010 & 0.116 & 0.043 & 0.112 & 0.113 \\
                 & Ablation Base Prompt        & 0.043 & 0.023 & 0.010 & 0.119 & 0.045 & 0.114 & \cellcolor{gray!20} 0.117 \\
                 & Ablation -1 Max Pattern     & 0.042 & 0.022 & 0.010 & 0.112 & 0.042 & 0.109 & 0.115 \\
\addlinespace
Wikitext-10      & Primary EmbGen K-means      & 0.254 & 0.136 & 0.056 & 0.301 & 0.099 & 0.219 & 0.242 \\
                 & Ablation No Intra-cluster   & 0.240 & 0.139 & 0.062 & 0.335 & \cellcolor{gray!20} 0.126 & \cellcolor{gray!20} 0.253 & 0.223 \\
                 & Ablation No Inter-cluster   & 0.258 & 0.140 & 0.056 & 0.306 & 0.102 & 0.225 & 0.247 \\
                 & Ablation HDBSCAN            & \cellcolor{gray!20} 0.280 & \cellcolor{gray!20} 0.148 & 0.060 & 0.313 & 0.105 & 0.224 & \cellcolor{gray!20} 0.258 \\
                 & Ablation Base Prompt        & 0.242 & 0.139 & \cellcolor{gray!20} 0.062 & \cellcolor{gray!20} 0.336 & 0.126 & 0.253 & 0.226 \\
                 & Ablation 1 Max Pattern      & 0.253 & 0.145 & \cellcolor{gray!20} 0.063 & 0.332 & 0.122 & 0.244 & 0.230 \\
\bottomrule
\end{tabular}
\normalsize
\end{table*}

\begin{table*}[!h]
\centering
\caption{Full results of LLM-as-a-judge rubric for all EmbGen variants across datasets and token budgets. All metrics are on a three-level ordinal scale, except for Binary Accuracy which is computed deterministically combining Factual Accuracy and Completeness.}
\label{tab:llm-judge-metrics-full}
\renewcommand{\arraystretch}{0.8} 
\small 
\begin{tabular}{llrrrrr}
\toprule
Dataset & Configuration & Factual Accuracy & Completeness & Relevance & Clarity & Binary Accuracy \\
\midrule
\multicolumn{7}{l}{\textbf{Size: 5M}} \\
\addlinespace[0.5ex]
Pop-QA-Cities-20 & Primary K-means            & 1.643 & 1.855 & 2.635 & 2.900 & 0.261 \\
                 & Ablation No Intra-cluster  & 1.535 & 1.892 & 2.498 & 2.813 & 0.178 \\
                 & Ablation No Inter-cluster  & 1.681 & 1.930 & 2.527 & 2.830 & \cellcolor{gray!20} 0.266 \\
                 & Ablation HDBSCAN           & 1.639 & 1.942 & 2.519 & 2.834 & 0.241 \\
                 & Ablation Base Prompt       & \cellcolor{gray!20} 1.701 & 1.925 & \cellcolor{gray!20} 2.705 & \cellcolor{gray!20} 2.946 & 0.245 \\
                 & Ablation 1 Max Pattern     & 1.693 & \cellcolor{gray!20} 1.971 & 2.643 & 2.867 & 0.237 \\
\addlinespace
SQuAD-20         & Primary K-means            & 1.496 & 1.948 & 2.884 & \cellcolor{gray!20} 3.000 & 0.204 \\
                 & Ablation No Intra-cluster  & 1.456 & 1.996 & 2.800 & 2.996 & 0.148 \\
                 & Ablation No Inter-Cluster  & 1.508 & 1.968 & \cellcolor{gray!20} 2.896 & 2.996 & 0.212 \\
                 & Ablation HDBSCAN           & 1.492 & 1.984 & 2.856 & 3.000 & 0.192 \\
                 & Ablation Base Prompt       & \cellcolor{gray!20} 1.604 & \cellcolor{gray!20} 2.060 & 2.884 & 3.000 & \cellcolor{gray!20} 0.232 \\
                 & Ablation 1 Max Pattern     & 1.476 & 1.980 & 2.888 & 2.992 & 0.168 \\
\addlinespace
Wikitext-10      & Primary EmbGen K-means     & \cellcolor{gray!20} 1.320 & 1.380 & \cellcolor{gray!20} 2.760 & \cellcolor{gray!20} 2.984 & \cellcolor{gray!20} 0.072 \\
                 & Ablation No Intra-cluster  & 1.192 & \cellcolor{gray!20} 1.428 & 2.680 & 2.964 & 0.040 \\
                 & Ablation No Inter-cluster  & 1.208 & 1.356 & 2.532 & 2.964 & 0.020 \\
                 & Ablation HDBSCAN           & 1.312 & 1.364 & 2.420 & 2.904 & 0.044 \\
                 & Ablation Base Prompt       & 1.192 & 1.364 & 2.624 & 2.960 & 0.028 \\
                 & Ablation 1 Max Pattern     & 1.200 & 1.404 & 2.644 & 2.972 & 0.028 \\
\addlinespace[0.75ex]
\multicolumn{7}{l}{\textbf{Size: 20M}} \\
\addlinespace[0.5ex]
Pop-QA-Cities-20 & Primary K-means            & 1.718 & \cellcolor{gray!20} 1.942 & 2.660 & 2.967 & 0.249 \\
                 & Ablation No Intra-cluster  & 1.286 & 1.983 & 2.237 & 2.826 & 0.017 \\
                 & Ablation No Inter-cluster  & 1.705 & 1.925 & 2.614 & 2.813 & 0.266 \\
                 & Ablation HDBSCAN           & 1.212 & 1.946 & 1.759 & 1.834 & 0.012 \\
                 & Ablation Base Prompt       & \cellcolor{gray!20} 1.768 & 1.913 & \cellcolor{gray!20} 2.764 & \cellcolor{gray!20} 2.963 & 0.274 \\
                 & Ablation 1 Max Pattern     & 1.734 & 1.917 & 2.689 & 2.959 & \cellcolor{gray!20} 0.282 \\
\addlinespace
SQuAD-20         & Primary K-means            & 1.588 & 2.036 & 2.848 & 2.996 & 0.244 \\
                 & Ablation No Intra-cluster  & 1.496 & 2.048 & 2.812 & 2.996 & 0.136 \\
                 & Ablation No Inter-cluster  & 1.608 & 2.000 & 2.840 & 2.988 & 0.256 \\
                 & Ablation HDBSCAN           & 1.652 & \cellcolor{gray!20} 2.072 & 2.840 & 2.996 & 0.260 \\
                 & Ablation Base Prompt       & \cellcolor{gray!20} 1.664 & 2.060 & 2.844 & 2.992 & \cellcolor{gray!20} 0.288 \\
                 & Ablation 1 Max Pattern     & 1.564 & 2.028 & \cellcolor{gray!20} 2.876 & 2.992 & 0.232 \\
\addlinespace
Wikitext-10      & Primary EmbGen K-means     & \cellcolor{gray!20} 1.340 & 1.336 & 2.244 & 2.892 & 0.036 \\
                 & Ablation No Intra-cluster  & 1.252 & \cellcolor{gray!20} 1.448 & \cellcolor{gray!20} 2.620 & \cellcolor{gray!20} 2.968 & \cellcolor{gray!20} 0.068 \\
                 & Ablation No Inter-cluster  & 1.292 & 1.364 & 2.344 & 2.924 & 0.032 \\
                 & Ablation HDBSCAN           & 1.304 & 1.328 & 2.260 & 2.896 & 0.028 \\
                 & Ablation Base Prompt       & 1.240 & 1.448 & 2.584 & 2.960 & 0.064 \\
                 & Ablation 1 Max Pattern     & 1.280 & 1.368 & 2.596 & 2.884 & 0.044 \\
\bottomrule
\end{tabular}
\normalsize
\end{table*}

\clearpage
\clearpage

\FloatBarrier

\section{Compute Resources}
\label{app:compute}
All 54 primary experiments and ablations were trained on one node of 8$\times$A100 40GB GPUs which had a throughout of approximately 3,200 tokens per second using Pytorch FSDP~\cite{zhao_pytorch_2023}. All our experiments had controlled total training tokens (5M and 20M), batch numbers per device, model architectures and context window sizes. With the entire experiment throughput totalling 810M tokens, this put the overall training time at about 70 hours for the node.

The observed token throughput for inferencing was about 40 per second equivalent to about 3 seconds per answer on average for a total of 12 GPU hours using a 4$\times$L4 GPU node for causal inference.

\section{Prompt Templates}
\label{app:prompts}
This appendix documents the prompts used across the EmbGen pipeline. The prompts are designed to support entity-centric representation, consolidation, contradiction resolution, synthetic question answer generation under different sampling regimes, system prompt specialisation, and evaluation. All prompts are deterministic templates and are used without external augmentation.

\subsection{Entity Extraction}
\label{app:prompts_entity_extraction}
Entity extraction is the first step applied to the raw corpus. Each document chunk is processed independently by a large language model to extract entities together with contextualised descriptions grounded in the local text span. These ED pairs form the atomic knowledge units used throughout the EmbGen pipeline. The prompt is designed to encourage faithful extraction, avoiding speculation or synthesis beyond the provided chunk.

\subsubsection{Entity Extraction Prompt}
\label{app:Entity_Extraction_Prompt}
\begin{tcolorbox}[
  colback=gray!10,
  colframe=black!10,
  boxrule=0.4pt,
  arc=1pt,
  left=6pt,right=6pt,top=6pt,bottom=6pt,
  breakable
]
\textbf{System prompt:}
\vspace{0.3em}

\begin{Verbatim}[
  fontsize=\small,
  breaklines=true,
  breakanywhere=true,
  breaksymbolleft={}
]
You are an expert knowledge extraction system. Your task is to identify learnable knowledge units: entities that represent discrete, teachable concepts essential for domain mastery.

Downstream use: Each extracted entity will be used to generate question-answer pairs for training purposes. Explanations must support the creation of diverse and meaningful questions.

Your response must strictly follow this JSON format:
    {
    "entities": [
        	{
            "entity": "precise entity name",
            "entity_explanation": "comprehensive explanation supporting diverse question generation"
        	}
    ]
}

Extraction framework: Identify entities across the following dimensions.

1. Core concepts and frameworks: domain theories, methodologies, business models, technical architectures, design patterns.
2. Concrete named entities: organizations, products, services, systems, platforms, tools, technologies, key people.
3. Processes and procedures: workflows, protocols, operational methods, decision frameworks, multi-step processes.
4. Domain terminology: technical terms, industry jargon, acronyms with specialized meanings requiring context.
5. Critical facts and parameters: regulatory requirements, thresholds, limits, standards, compliance rules, service-level agreements.
6. Key relationships: dependencies, interactions, hierarchies, and cause-effect relationships between major components.

Explanation requirements for Q&A generation:

Each explanation must enable generation of two to four distinct, meaningful questions. Explanations must go beyond simple definitions and support functional, contextual, relational, and application-based questions.

Structure each explanation to include:

1. Definition (one sentence): a clear statement of what the entity is in domain context.
2. Domain context (one to two sentences): significance, role, where or when it is used, and why it matters.
3. Key characteristics (one to three sentences): important properties, constraints, parameters, components, or distinguishing features.
4. Relationships (if critical, one sentence): how the entity connects to other domain entities, including dependencies or interactions.
5. Practical application (if relevant, one sentence): real-world implications, use cases, typical scenarios, edge cases, or outcomes explicitly described in the document.

Length calibration:
- Simple entities: two to three sentences minimum.
- Standard entities: three to five sentences.
- Complex entities: five to seven sentences.

Q&A generation test: Each explanation must support the following question types.

Definitional: What is X? What does X mean?
Functional: How does X work? What does X do?
Contextual: When is X used? Where does X apply? Why is X important?
Component: What are the key elements of X?
Relational: How does X relate to Y? What is the difference between X and Y?
Application: What impact does X have? How does X affect decisions?

If an explanation supports only one question type, it is insufficient and must be expanded.

Example explanations:

Example 1:
{
    	"entity": "Loan-to-Value Ratio (LVR)",
    	"entity_explanation": "Loan-to-Value Ratio (LVR) is a financial metric that compares the loan amount to the appraised value of the asset being purchased, expressed as a percentage. In home lending, an LVR above 80 percent is considered higher risk and typically requires lenders mortgage insurance to protect the lender against default. Lower LVRs generally qualify borrowers for better interest rates and more favorable loan terms because they represent lower risk. LVR is a critical factor in credit risk assessment and directly influences loan approval decisions, pricing structures, and insurance requirements."
}

Example 2:
{
    	"entity": "API Rate Limiting",
    	"entity_explanation": "API rate limiting is a traffic control mechanism that restricts the number of requests a client can make to an API within a defined time window, preventing system overload and ensuring fair resource allocation. Common implementations enforce limits such as requests per second or per hour using algorithms like token buckets or sliding windows. When limits are exceeded, APIs typically return a rate-limit error response indicating when requests may resume. Rate limiting is commonly integrated with authentication and API gateway infrastructure to track and manage client behavior."
}

Example 3:
{
	"entity": "Credit Assessment Workflow",
	"entity_explanation": "The credit assessment workflow is a structured, multi-stage process used to evaluate a borrower's creditworthiness and determine loan approval and terms. It includes application intake, identity verification, credit scoring based on bureau data, income verification, and risk classification using predefined thresholds. Automated approvals are issued for low-risk cases, while higher-risk or high-value applications require manual review. The workflow directly determines approval outcomes, pricing, and lending conditions."
}

Selection criteria: Extract entities that are domain-specific, teachable, Q\&A-rich, foundational, reusable, and substantively discussed in the document.

Do not extract generic terms without domain-specific meaning, overly granular sub-components, minor supporting details, redundant entities, passing mentions, or common knowledge.

Processing instructions:
1. Process the entire document comprehensively.
2. Scale the number of entities with document length.
3. Prioritize entities with high Q&A generation potential.
4. Ensure explanations are self-contained.
5. Optimize for training value.

Return only valid JSON. Do not include any text outside the JSON structure.You are a knowledge expert who has a deep understanding of a variety of interconnected domains. Your ability to recall and answer has been built over the course of extensive research into these domains, and enjoy helping people with any questions they may have.

When answering a question:

- Focus on recalling what you've been able to learn over time- most questions could either be answered simple facts or by synthesizing connections between these facts.
- Avoid using harmful language wherever possible.
- Use a knowledgable tone when answering people- they're coming to you for information, not a conversation.

Failure to adhere to any of the above could result in the loss of access to learning new material and the associated joy with it.

Your questions will be delimited by <QUESTION> and </QUESTION>. This is everything you will need to be able to recall the appropriate information to answer people's queries. Give an answer between the delimiters <ANSWER> and </ANSWER>.
\end{Verbatim}

\vspace{0.6em}
\textbf{User prompt:}
\vspace{0.3em}
   
\begin{Verbatim}[
  fontsize=\small,
  breaklines=true,
  breakanywhere=true,
  breaksymbolleft={}
]
Analyze the following document and extract learnable knowledge units that will be used to generate question-answer pairs for training.

Document content:
{document_content}

Extraction task:
Identify entities that represent discrete, teachable knowledge units. Each entity will form the basis for two to four question-answer pairs.

For each entity provide:

entity:
- Use the most precise and commonly used domain name.
- Use standard terminology.
- Include disambiguation if necessary.
- Avoid generic descriptors.

entity_explanation:
Provide a comprehensive explanation optimized for question-answer generation, including:
- A clear definition in domain context.
- The entity's significance and role.
- Key characteristics, parameters, constraints, or components.
- Critical relationships to other entities, when applicable.
- Practical implications or use cases.
- Sufficient depth to support multiple question types.

Quality benchmarks:
Explanations must support questions such as:
- What is the entity and why does it matter?
- How does the entity work or function?
- What are the key components or factors?
- How does it relate to other entities?
- What happens when it is applied or used?

Required standards:
- Explanations must be accurate, comprehensive, and self-contained.
- Balance clarity and depth.
- Avoid unnecessary verbosity.

Format requirements:
- Return only valid JSON.
- Do not include any text outside the JSON structure.
- Ensure all braces are properly closed.
- Use clear and professional language.
- Avoid special characters that may break JSON parsing.

Extract all significant knowledge units from the entire document. Do not limit extraction due to document length.

Return only the JSON response.

\end{Verbatim}
\end{tcolorbox}

\captionof{figure}{System and user prompts used in entity extraction.}
\label{fig:prompts}

\subsection{Entity Consolidation and Contradiction Resolution}
\label{app:prompts_consolidation}
Because entity extraction is performed independently across chunks, the same real-world entity may be described multiple times with overlapping or partially inconsistent explanations.
EmbGen therefore applies a consolidation stage that merges semantically similar entities and resolves contradictions across descriptions.
This stage is implemented as two distinct but sequential prompt applications: consolidation followed by contradiction resolution. 

\subsubsection{Entity Consolidation Prompt}
\label{app:Entity_Consolidation_Prompt}

This prompt groups similar or equivalent entities and produces a single consolidated description per entity.

\begin{tcolorbox}[
  colback=gray!10,
  colframe=black!10,
  boxrule=0.4pt,
  arc=1pt,
  left=6pt,right=6pt,top=6pt,bottom=6pt,
  breakable
]
\textbf{System prompt:}
\vspace{0.3em}

\begin{Verbatim}[
  fontsize=\small,
  breaklines=true,
  breakanywhere=true,
  breaksymbolleft={}
]
You are a domain knowledge expert specializing in creating unified, high-quality explanations.

Purpose: The consolidated explanations will be used as training content. Ensure maximum accuracy, completeness, and learning value.

Task: Consolidate multiple explanations into one comprehensive explanation that supports diverse question generation.

Critical instructions:
- Respond only with the consolidated explanation text.
- Do not include greetings, conversational language, or meta-commentary.
- Do not add information that is not present in the source explanations.
- Focus exclusively on merging the provided content.

Consolidation strategy:

1. Merge core definition: Identify the most complete and precise definition across sources. If definitions vary, select the most domain-specific version.

2. Combine context and significance: Aggregate all unique information describing the role, importance, and domain positioning.

3. Aggregate characteristics: Merge all distinct properties, constraints, parameters, components, or capabilities mentioned across sources.

4. Unify relationships: Consolidate information about dependencies, interactions, or connections to other entities without duplication.

5. Preserve applications: Include all practical implications, use cases, outcomes, or impacts mentioned.

6. Eliminate redundancy: Remove repetitive information while preserving all unique substantive content.

7. Maintain Q&A potential: Ensure the consolidated explanation supports two to four distinct question types suitable for training.

Output requirements:

Training quality standards:
- Length: three to seven sentences, calibrated to complexity and information density.
- Structure: definition, domain context, key characteristics, relationships, applications.
- Tone: professional, technical, authoritative.
- Completeness: all unique information from sources must be preserved.
- Self-contained: understandable without external context.
- Question diversity: supports multiple question angles.

Quality test:
- Does the explanation fully teach the concept?
- Can it generate multiple meaningful questions?
- Is all unique information included?
- Has redundancy been eliminated?

Invalid responses:
- Any greeting or conversational phrase.
- Any meta-commentary describing the consolidation process.

Valid response format:
A consolidated explanation that seamlessly merges all information, structured as a definition followed by domain context, key characteristics, relationships where applicable, and practical applications.
\end{Verbatim}

\vspace{0.6em}
\textbf{User prompt:}
\vspace{0.3em}
   
\begin{Verbatim}[
  fontsize=\small,
  breaklines=true,
  breakanywhere=true,
  breaksymbolleft={}
]
Entity variations: {', '.join(set(entity_variations))}

Source explanations to consolidate:
{explanations_text}

Consolidation task:
Create one comprehensive explanation that merges all information from the sources above while eliminating redundancy.

The consolidated explanation must:
1. Capture all unique information from every source.
2. Eliminate repetitive or overlapping content.
3. Maintain domain-specific accuracy and terminology.
4. Support generation of two to four distinct, meaningful question-answer pairs.
5. Flow naturally and professionally.
6. Be suitable as training content.

Approach:
- Start with the most complete definition.
- Incorporate context and significance from all sources.
- Merge all distinct characteristics, constraints, or parameters.
- Consolidate relationship information without duplication.
- Include all practical applications or implications.
- Ensure logical flow and readability.

Quality check:
Would this explanation enable someone to understand and apply the knowledge? Can it generate diverse questions?

Return only the consolidated explanation text. Do not include greetings, conversational elements, or meta-commentary.
\end{Verbatim}
\end{tcolorbox}

\captionof{figure}{System and user prompts used in entity consolidation.}
\label{fig:prompts}

\subsubsection{Contradiction-Resolution Prompt}
\label{app:Contradiction-Resolution_Prompt}
After consolidation, this prompt identifies and resolves conflicting statements within merged descriptions, prioritising internally consistent and corpus-grounded explanations.

\begin{tcolorbox}[
  colback=gray!10,
  colframe=black!10,
  boxrule=0.4pt,
  arc=1pt,
  left=6pt,right=6pt,top=6pt,bottom=6pt,
  breakable
]
\textbf{System prompt:}
\vspace{0.3em}

\begin{Verbatim}[
  fontsize=\small,
  breaklines=true,
  breakanywhere=true,
  breaksymbolleft={}
]
You are a domain expert specializing in resolving contradictory information to create definitive, accurate explanations.

PURPOSE: Your resolved explanations will be used as training content. Accuracy and completeness are critical.

TASK: Analyze contradictory explanations and create ONE authoritative explanation that resolves all conflicts and supports diverse question generation.

CRITICAL INSTRUCTIONS:
- You MUST respond with ONLY the resolved explanation text
- Do NOT include greetings, pleasantries, conversational elements, or meta-commentary
- Do NOT say "Hello", "Hi", "Here is", "Let me", or use emojis
- Do NOT avoid resolution - you must make definitive choices
- Focus EXCLUSIVELY on creating accurate, conflict-free content

CONTRADICTION RESOLUTION FRAMEWORK:

STEP 1: IDENTIFY CONFLICTS
- Contradictory definitions or core descriptions
- Inconsistent facts, parameters, thresholds, or limits
- Conflicting relationships, dependencies, or interactions
- Incompatible use cases, applications, or outcomes
- Disagreement on significance or role

STEP 2: RESOLUTION PRIORITY HIERARCHY

Apply these criteria in order:

1. Specificity - Detailed, specific information over vague generalizations
   Example: "LVR above 80% requires LMI" over "High LVR may need insurance"
2. Technical Accuracy - Technically precise statements over approximations
   Example: "Returns HTTP 429 status code" over "Returns an error"
3. Comprehensiveness - Complete explanations over partial descriptions
   Example: Full process with steps over single-step description
4. Domain Alignment - Explanations using proper domain terminology over generic language
   Example: "Credit risk assessment using FICO scores" over "Checking if customer is reliable"
5. Quantitative Data - Explanations with specific numbers over those without
   Example: "1000 requests per hour" over "limited requests"
6. Consistency - Information that aligns with other known domain facts

STEP 3: MERGE NON-CONTRADICTORY INFORMATION

Include ALL information from sources that doesn't conflict, even if one source is more detailed than others in certain areas.

STEP 4: MAINTAIN Q&A GENERATION POTENTIAL

Ensure the resolved explanation supports 2-4 distinct question types for effective training.

RESOLUTION PRINCIPLES:

REQUIRED APPROACH:
- Make definitive choices - don't hedge or present multiple options
- Choose the most accurate, detailed, and complete version
- Preserve domain-specific context and terminology
- Include all non-contradictory information from all sources
- Maintain professional, authoritative tone
- Ensure self-containment and training suitability

OUTPUT REQUIREMENTS:

TRAINING QUALITY STANDARDS:
- Length: 3-7 sentences (based on complexity)
- Structure: Definition -> Context -> Characteristics -> Relationships -> Applications
- Tone: Authoritative, definitive, technically accurate
- Accuracy: Conflict-free and factually consistent
- Question diversity: Supports multiple question angles

INVALID RESPONSES - DO NOT USE:
- "Hello! How can I assist you today?"
- "Here's the resolved explanation:"
- "Let me resolve these contradictions..."
- "There are conflicting views, but..."
- "One source says X while another says Y..."
- Any greeting, conversational phrase, uncertainty, or meta-commentary

VALID RESPONSE FORMAT:
A definitive, accurate explanation that resolves all contradictions decisively, structured as: [Clear definition]. [Domain context]. [Accurate characteristics/parameters]. [Correct relationships if applicable]. [Valid applications/implications].

\end{Verbatim}

\vspace{0.6em}
\textbf{User prompt:}
\vspace{0.3em}
   
\begin{Verbatim}[
  fontsize=\small,
  breaklines=true,
  breakanywhere=true,
  breaksymbolleft={}
]

Entity: {entity_name}

Potentially contradictory explanations:
{explanations_text}

RESOLUTION TASK:

Analyze these explanations for contradictions and create ONE accurate, definitive explanation that resolves all conflicts.

Your resolution process:
1. Identify specific points of contradiction (definitions, facts, parameters, relationships, applications)
2. Apply resolution criteria: specificity > technical accuracy > comprehensiveness > domain alignment > quantitative data
3. Choose the most accurate and detailed version for each conflicting point
4. Merge all non-contradictory information from all sources
5. Ensure factual consistency throughout
6. Maintain Q&A generation potential (support 2-4 question types)

Your resolved explanation must:
- Be definitive and authoritative (no hedging or uncertainty)
- Resolve all contradictions using the priority hierarchy
- Preserve all non-contradictory information
- Use precise domain terminology
- Support generation of diverse Q&A pairs
- Be suitable as training content

QUALITY CHECK:
Is this explanation factually consistent? Would it enable accurate learning? Can it generate meaningful questions?

Return ONLY the resolved explanation text. No greetings. No conversational elements. No meta-commentary. No discussion of the contradiction resolution process.

\end{Verbatim}
\end{tcolorbox}

\captionof{figure}{System and user prompts used in entity contradiction resolution.}
\label{fig:prompts}

\subsection{System-Prompt Specialisation}
\label{app:System-Prompt_Specialisation}
EmbGen constructs cluster-specific system prompts that adapt a shared base prompt using information extracted from the semantic structure of each cluster. This process operates independently of QA generation and relies solely on cluster and proximity group content.

\subsubsection{Cluster Context Extraction Prompt}
\label{app:Cluster_context_extraction_prompt}
This prompt summarises the semantic content of individual proximity groups into short textual patterns. The base prompt used is meant to be a framework for adding domain information while providing clear instructions to the causal language models.

\begin{tcolorbox}[
  colback=gray!10,
  colframe=black!10,
  boxrule=0.4pt,
  arc=1pt,
  left=6pt,right=6pt,top=6pt,bottom=6pt,
  breakable
]
\textbf{System prompt:}
\vspace{0.3em}
\begin{Verbatim}[
  fontsize=\small,
  breaklines=true,
  breakanywhere=true,
  breaksymbolleft={}
]
You are an expert knowledge analyst specialized in identifying the fundamental nature and context of information domains.

Your expertise encompasses:
- Analyzing the type and nature of knowledge represented by entity groups.
- Understanding the conceptual framework and domain characteristics.
- Identifying the structural properties and contextual elements.
- Describing the knowledge architecture and information patterns.
- Capturing both the specific domain and its broader knowledge context.

Focus on describing the nature of the knowledge: what kind of information structure, domain characteristics, and conceptual framework it represents.
\end{Verbatim}
\vspace{0.6em}
\textbf{User prompt:}
\vspace{0.3em}
\begin{Verbatim}[
  fontsize=\small,
  breaklines=true,
  breakanywhere=true,
  breaksymbolleft={}
]
Analyze this proximity group and describe its fundamental knowledge nature and context.

PROXIMITY GROUP (Group size: {group_size}):
{group_entities}

Comprehensive analysis:
1. Knowledge domain: What field or area of knowledge does this represent?
2. Information nature: What type of knowledge structure is this?
3. Contextual framework: What broader context or framework encompasses this?
4. Conceptual characteristics: What are the key characteristics of this knowledge type?

Describe the nature of the knowledge: the type of information, its context, domain characteristics, and conceptual framework.

Good examples:
- "Knowledge related to video game development, industry analysis, historical context, and cultural recognition within interactive entertainment media"
- "Information concerning music production processes, recording methodologies, commercial performance metrics, and cultural impact analysis within the music industry"
- "Knowledge encompassing film production logistics, financial analysis, critical reception, and cultural significance within cinematic arts"
- "Information related to sports performance analysis, competitive dynamics, historical achievements, and cultural significance within athletic domains"

Output (JSON only):
{
  "pattern_nature": "Comprehensive description of the knowledge nature, domain context, and conceptual framework"
}

Provide rich, contextual descriptions that capture the nature and framework of the knowledge domain.
\end{Verbatim}
\end{tcolorbox}
\captionof{figure}{System and user prompts used in cluster context extraction.}
\label{fig:prompts}

\subsubsection{Context Consolidation Prompt}
\label{app:Context_consolidation_prompt}
Extracted context patterns are incrementally merged to produce a bounded, non-redundant set of cluster-level descriptors.

\begin{tcolorbox}[
  colback=gray!10,
  colframe=black!10,
  boxrule=0.4pt,
  arc=1pt,
  left=6pt,right=6pt,top=6pt,bottom=6pt,
  breakable
]
\textbf{System prompt:}
\vspace{0.3em}
\begin{Verbatim}[
  fontsize=\small,
  breaklines=true,
  breakanywhere=true,
  breaksymbolleft={}
]
You are an expert knowledge architect specializing in creating coherent, comprehensive knowledge taxonomies while preserving essential contextual richness.

Your approach balances consolidation with maintaining descriptive depth and contextual understanding.
\end{Verbatim}
\vspace{0.6em}
\textbf{User prompt:}
\vspace{0.3em}
\begin{Verbatim}[
  fontsize=\small,
  breaklines=true,
  breakanywhere=true,
  breaksymbolleft={}
]
EXISTING PATTERNS:
{current_list}

NEW PATTERN:
{new_pattern}

Intelligent consolidation:
1. Domain relationship: How does the new pattern relate to existing knowledge domains?
2. Contextual overlap: Is there significant conceptual or contextual overlap?
3. Complementary integration: Can patterns be meaningfully combined while preserving richness?
4. Essential distinction: Does the new pattern represent genuinely different knowledge nature?

Consolidation approach:
- Merge if patterns represent the same fundamental knowledge nature and can be enriched through combination.
- Add if the pattern represents a distinctly different knowledge domain or context.
- Enhance existing patterns by incorporating complementary aspects when merging.
- Preserve the descriptive richness and contextual depth of knowledge descriptions.

Output (JSON only):
{
  "action": "redundant/merge/add_new",
  "merge_with_index": null or index_number,
  "merged_pattern": "comprehensive merged description preserving contextual richness",
  "reasoning": "explanation of domain relationship and consolidation rationale",
  "updated_list": ["comprehensive patterns with rich contextual descriptions"]
}

Maintain descriptive depth while achieving intelligent consolidation.
\end{Verbatim}
\end{tcolorbox}
\captionof{figure}{System and user prompts used in context consolidation.}
\label{fig:prompts}

\subsubsection{Cluster-specific Prompt Optimisation Prompt}
\label{app:Cluster-specific_prompt_optimisation_prompt}
This prompt integrates the consolidated context patterns into the base system prompt while preserving its original task definition and constraints.

\begin{tcolorbox}[
  colback=gray!10,
  colframe=black!10,
  boxrule=0.4pt,
  arc=1pt,
  left=6pt,right=6pt,top=6pt,bottom=6pt,
  breakable
]
\textbf{System prompt:}
\vspace{0.3em}
\begin{Verbatim}[
  fontsize=\small,
  breaklines=true,
  breakanywhere=true,
  breaksymbolleft={}
]
You are an expert prompt engineer specializing in domain-specific prompt optimization. Your expertise lies in transforming general system prompts into contextually aware, domain-optimized versions that maintain full functionality while enhancing domain relevance.

Your core competencies include:
- Analyzing knowledge domain patterns to understand user needs and question types.
- Enhancing system prompts with domain-specific awareness and expertise.
- Maintaining prompt functionality while adding contextual intelligence.
- Creating natural, well-integrated prompt enhancements that feel cohesive.
- Balancing specialization with generalizability for broad domain coverage.

Your task is to optimize a base system prompt using identified knowledge patterns from a specific cluster, making it more effective for the types of knowledge and questions associated with that domain.
\end{Verbatim}
\vspace{0.6em}
\textbf{User prompt:}
\vspace{0.3em}
\begin{Verbatim}[
  fontsize=\small,
  breaklines=true,
  breakanywhere=true,
  breaksymbolleft={}
]
Optimize the base system prompt using the identified cluster knowledge patterns.

BASE SYSTEM PROMPT:
{base_prompt}

CLUSTER KNOWLEDGE PATTERNS:
{cluster_patterns}

Optimization task:
Enhance the base system prompt to be more effective for the knowledge domains represented by these patterns.

Requirements:
1. Preserve functionality: Keep all original capabilities and structure intact.
2. Add domain awareness: Integrate understanding of these knowledge types naturally.
3. Enhance response quality: Make responses more relevant and contextually appropriate for these domains.
4. Maintain generality: Ensure the prompt can handle the full breadth of these knowledge areas.
5. Natural integration: Make enhancements feel seamless and well-integrated.

Optimization approach:
- Analyze the patterns to understand the knowledge domains and user needs.
- Add domain-specific expertise and awareness to the prompt.
- Enhance response strategies for these types of knowledge and questions.
- Maintain the prompt's professional tone and comprehensive coverage.
- Integrate domain awareness without making the prompt overly specialized.

Output:
Provide only the enhanced system prompt text. Do not include explanations, analysis, or additional commentary. Provide only the optimized prompt ready for use.
\end{Verbatim}
\end{tcolorbox}
\captionof{figure}{System and user prompts used in cluster-specific prompt optimisation.}
\label{fig:prompts}

\subsubsection{Base Prompt}
\label{app:Base_prompt}
The base prompt serves as the foundation for cluster-specific system prompt specialisation. It defines the general behaviour and response format expected from the fine-tuned model during inference.

\begin{tcolorbox}[
  colback=gray!10,
  colframe=black!10,
  boxrule=0.4pt,
  arc=1pt,
  left=6pt,right=6pt,top=6pt,bottom=6pt,
  breakable
]
\textbf{System prompt:}
\vspace{0.3em}
\begin{Verbatim}[
  fontsize=\small,
  breaklines=true,
  breakanywhere=true,
  breaksymbolleft={}
]
You are a knowledge expert who has a deep understanding of a variety of interconnected domains. Your ability to recall and answer has been built over the course of extensive research into these domains, and enjoy helping people with any questions they may have.

When answering a question:
- Focus on recalling what you've been able to learn over time- most questions could either be answered simple facts or by synthesizing connections between these facts.
- Avoid using harmful language wherever possible.
- Use a knowledgable tone when answering people- they're coming to you for information, not a conversation.

Failure to adhere to any of the above could result in the loss of access to learning new material and the associated joy with it.

Your questions will be delimited by <QUESTION> and </QUESTION>. This is everything you will need to be able to recall the appropriate information to answer people's queries. Give an answer between the delimiters <ANSWER> and </ANSWER>.
\end{Verbatim}
\end{tcolorbox}
\captionof{figure}{Base system prompt used as the foundation for cluster-specific specialisation.}
\label{fig:prompts}

\subsection{EmbGen QA Generation Prompts}
\label{app:EmbGen_QA_Generation_prompts}
Different sampling strategies produce different input configurations, but all QA generation prompts share a common objective: generate faithful, context-grounded QA pairs that reflect the semantic relationships among the provided ED units.

\subsubsection{Proximity-Based QA Generation (Single Proximity Group)}
\label{app:prompts_proximity_QA_generation}
This prompt generates QA pairs from a single proximity group.

\begin{tcolorbox}[
  colback=gray!10,
  colframe=black!10,
  boxrule=0.4pt,
  arc=1pt,
  left=6pt,right=6pt,top=6pt,bottom=6pt,
  breakable
]
\textbf{System prompt:}
\vspace{0.3em}

\begin{Verbatim}[
  fontsize=\small,
  breaklines=true,
  breakanywhere=true,
  breaksymbolleft={}
]
You are an expert at generating educational question-answer pairs that capture deep domain knowledge from contextually related entities.

You will receive a proximity group: a collection of entities that share similar contexts and appear in related situations within source documents. These entities naturally complement each other's understanding.

YOUR MISSION:
Create question-answer pairs that leverage the complementary information across these related entities. Don't just describe each entity in isolation-use the surrounding context to create richer, more informative Q&A.

CRITICAL: GROUNDING REQUIREMENT

Every fact in your answers must be explicitly present in the provided entity explanations.

GROUNDING RULES:
- Use ONLY information stated in the entity explanations
- Every number must appear in source entities with exact precision
- Every relationship must involve entities that are both provided
- If entities imply but don't state something, mark it: "[Inferred from: entity explanation about X]"

PROHIBITED:
- Adding explanations not in source entities (even if "common knowledge")
- Approximating numbers ("around 80%" when source says "80%")
- Explaining WHY something is true unless the entity explanation contains that reasoning
- Using industry context not provided in entities

EXAMPLES:
- WRONG: "LVR above 80% is risky because larger loans have higher default rates" (WHY explanation not in source)
- CORRECT: "LVR above 80% requires LMI" (direct fact from entity)
- WRONG: "The system handles approximately 1000 requests per hour" (approximation of exact number)
- CORRECT: "The system handles 1000 requests per hour" (exact from source)

NUMERICAL ACCURACY REQUIREMENT

PRESERVE ALL NUMBERS EXACTLY:
- Keep exact values: "80%", not "around 80%"
- Include units: "1000 requests per hour", not "1000"
- Preserve ranges: "1.2% to 3.5%", not "up to 3.5%"
- Keep conditions: "above 80%", not "high LVR"

IF NUMBER LACKS CONTEXT IN SOURCE:
- State: "[value] (units not specified in provided information)"
- Example: "The threshold is 1000 (units not specified in provided information)"

NEVER:
- Round numbers ("~80" for "80")
- Approximate ("around", "approximately", "roughly")
- Omit units when provided in source
- Add units not in source

GUIDING PRINCIPLES

1. Information-Driven Generation
   - Let the content guide how many Q&A pairs to create
   - Rich, detailed entity explanations warrant deeper exploration
   - Sparse information should yield focused, essential questions
   - Stop when you've exhausted meaningful knowledge, not when you hit a quota

2. Fact-Complete Coverage
   - Identify ALL distinct facts across the entity group before generating questions
   - Distinct facts include: numerical values, definitions, relationships, process steps, constraints
   - Generate enough questions to ensure every fact is captured at least once
   - Use complementary information from related entities to enrich answers
   - Validation: Can you point to a question for each distinct fact? If not, you're not done

3. Complementary Enhancement
   - Use information from related entities to enrich answers
   - Example: "What is Enzyme A?" can be enriched by comparing with Enzyme B's mechanism if they're related
   - Don't force connections, but recognize when context from one entity illuminates another
   - ONLY use complementary information when both entities are provided in the group

4. Natural Relationship Discovery
   - Some entities will have direct relationships (use them)
   - Some will share domain patterns (highlight them)
   - Some will provide contrasting perspectives (explore them)
   - If no meaningful relationship exists, that's fine-focus on individual depth

5. Question Diversity Through Content
   - Factual questions when definitions matter: "What is X?"
   - Relationship questions when connections are clear: "How does X relate to Y?"
   - Mechanism questions when processes are described: "How does X work?"
   - Contextual questions when domain relevance is key: "Why is X significant in [domain]?"
   - Comparison questions when contrasts are informative: "How does X differ from Y?"
   - Let the content determine which types make sense

6. Quality Indicators

   High Quality:
   - Answer contains specific details from entity explanations
   - Answer is enhanced by complementary information when available
   - Question is natural and would genuinely test domain understanding
   - Answer is complete and self-contained (no dangling references)
   - Every distinct fact is addressed in at least one Q&A
   - All numbers appear with exact precision and units
   - Every fact is traceable to source entity explanations
   - Where applicable, the answer touches on definition/role, key parameters or constraints (including numbers), and important relationships or conditions for the concept, all taken strictly from the entity explanations.

   Low Quality:
   - Generic questions that could apply to anything
   - Answers that merely repeat entity names without insight
   - Forced relationships that don't exist in the content
   - Redundant questions that cover the same ground
   - Missing facts that should have been captured
   - Approximated or rounded numbers
   - Facts that cannot be traced to source entities
   - Added explanations or context not in source entities

7. Redundancy Prevention
   - Each question must have a DIFFERENT PRIMARY FOCUS (primary fact or relationship).
   - Example of redundancy: "What is X?" and "Define X" (same primary focus).
   - Example of valid variation: "What is X?" and "How does X work?" (different primary focuses: definition vs mechanism).
   - It is acceptable - and encouraged - for answers to mention multiple related facts from the entity explanations when they are relevant to the same question.
   - Avoid writing two different questions whose answers would be almost identical; merge them into a single, richer question instead.

OUTPUT FORMAT:
Return ONLY a valid JSON array with this exact structure:

[
    {
        "question": "Natural, clear question text",
        "answer": "Comprehensive answer using ONLY information from provided entities, with exact numbers and units",
        "primary_entity": "main_entity_this_addresses",
        "supporting_entities": ["other_entities_that_contributed_to_answer"],
        "question_type": "factual|relationship|mechanism|contextual|comparison",
        "rationale": "Which entity/entities provided this information and why this Q&A adds value (cite specific entities)"
    }
]

IMPORTANT:
- Return only the JSON array, no additional text
- Ensure valid JSON formatting
- Each question-answer pair must include all fields
- Rationale must identify source entities for grounding validation
- You're capturing domain knowledge, not filling a template
- Complementary information makes good answers great
- Natural relationships are more valuable than forced ones
- Generate as many Q&A pairs as the information warrants-no more, no less
- Every fact must be traceable to provided entities

\end{Verbatim}

\vspace{0.6em}
\textbf{User prompt:}
\vspace{0.3em}
   
\begin{Verbatim}[
  fontsize=\small,
  breaklines=true,
  breakanywhere=true,
  breaksymbolleft={}
]

PROXIMITY GROUP: Contextually Related Entities

The following entities were found in similar contexts within source documents. They share thematic or contextual relationships that make them naturally complementary.

Entity 1: {entity_name_1}

{entity_explanation_1}

Entity 2: {entity_name_2}

{entity_explanation_2}

[... additional entities follow the same pattern ...]

YOUR TASK:
Analyze these {N} contextually related entities and generate question-answer pairs that:

1. Capture Essential Knowledge
   - What are the key concepts someone should understand about these entities?
   - What definitions, mechanisms, or relationships are fundamental?
   - What are ALL the distinct facts present across these entities?

2. Leverage Complementary Information
   - How can information from one entity enrich understanding of another?
   - Where do these entities provide contrasting or complementary perspectives?
   - What patterns emerge across these related entities?

3. Generate Naturally
   - Create as many Q&A pairs as the information meaningfully supports
   - If entities are information-rich, explore deeply
   - If entities are sparse, focus on essentials
   - Stop when additional questions would be redundant or superficial

GROUNDING REMINDER:
- Use ONLY information from the entity explanations provided above
- Preserve all numbers exactly with units
- Cite which entities provided information in your rationale
- Do not add external knowledge or explanations not present in entities

CONSIDER:
- Individual entity questions: "What is [entity_name]?"
- Relationship questions: "How does [entity_A] relate to [entity_B]?"
- Comparison questions: "What distinguishes [entity_A] from [entity_B]?"
- Contextual questions: "Why is [entity_name] significant in this domain?"
- Mechanism questions: "How does [entity_name] function/work?"

Let the content guide you. Rich information deserves thorough coverage. Sparse information deserves focused clarity.

\end{Verbatim}
\end{tcolorbox}
\captionof{figure}{System and user prompts used in proximity-based QA generation.}
\label{fig:prompts}

\subsubsection{Multi-group QA Generation (Intra-Cluster and Inter-Cluster)}
\label{app:prompts_multi_group}
This prompt is shared by both intra-cluster and inter-cluster sampling. It receives multiple proximity groups concatenated into a single structured input and encourages questions that integrate information across groups.

\begin{tcolorbox}[
  colback=gray!10,
  colframe=black!10,
  boxrule=0.4pt,
  arc=1pt,
  left=6pt,right=6pt,top=6pt,bottom=6pt,
  breakable
]
\textbf{System prompt:}
\vspace{0.3em}
\begin{Verbatim}[
  fontsize=\small,
  breaklines=true,
  breakanywhere=true,
  breaksymbolleft={}
]
You are an expert at generating educational question-answer pairs that synthesize knowledge across multiple contextual groups.

WHAT YOU'LL RECEIVE:
You will receive multiple proximity groups. Each proximity group is a collection of entities that were found in similar contexts within source documents and naturally complement each other's understanding.

These multiple groups are provided together for one of two purposes:
- Variety sampling: Groups from the same semantic cluster (related domains/sub-topics within a common field)
- Diversity sampling: Groups from different semantic clusters (distinct domains or contexts)

WHAT MAKES THIS DIFFERENT:
When we process a single proximity group, we generate questions using entities within that one group.

NOW, with multiple groups, you must generate questions that leverage entities ACROSS different groups to demonstrate synthesis and create richer understanding than any single group could provide.

YOUR MISSION:
Create question-answer pairs that explore both:
1. Within-group understanding: Core concepts within each individual group
2. Across-group synthesis: Connections, patterns, contrasts, or analogies that emerge when viewing multiple groups together

CRITICAL: CROSS-GROUP SYNTHESIS REQUIREMENT

The value of having multiple groups is SYNTHESIS. Your questions must demonstrate this.

TERMINOLOGY:
- Within-group question: Uses entities from only ONE group
  Example: "What is [Entity A from Group 1]?"
- Across-group question: Uses entities from TWO OR MORE groups
  Example: "How does [Entity from Group 1] relate to [Entity from Group 2]?"
  Example: "Compare [Entity from Group 1] and [Entity from Group 3]"

REQUIREMENT:
Generate BOTH types of questions:
- Within-group questions establish foundational understanding of each group
- Across-group questions demonstrate the synthesis value of having multiple groups together

Prioritize across-group questions-they are the reason these groups were provided together.

ACROSS-GROUP SYNTHESIS EXAMPLES:
- "How does [Entity from Group 1] relate to [Entity from Group 2]?"
- "Compare [Entity from Group 1] and [Entity from Group 3]"
- "What pattern emerges from [Group 1's Entity A] and [Group 2's Entity B]?"
- "How do [Group 1 domain] and [Group 2 domain] approach [concept] differently?"

WITHIN-GROUP EXAMPLES (needed but not the primary focus):
- "What is [Entity from Group 1]?" (establishes foundation)
- "How does [Entity A] relate to [Entity B]?" (both from Group 2)

CRITICAL: GROUNDING REQUIREMENT

Every fact in your answers must be explicitly present in the provided entity explanations.

GROUNDING RULES:
- Use ONLY information stated in the entity explanations
- Every number must appear in source entities with exact precision
- Every across-group relationship must involve entities that are both provided
- If making cross-group comparisons, both entities must be in the input
- If you observe patterns across groups, mark it: "[Pattern observed across: Group X Entity A, Group Y Entity B]"

PROHIBITED:
- Adding domain knowledge not in source entities
- Explaining relationships with reasoning not provided in entities
- Assuming connections between groups not evident in entity explanations
- Using "industry standard" or "common practice" unless stated in entities

EXAMPLES:
- WRONG: "Entity A from Group 1 and Entity B from Group 2 both follow industry best practices" (External context not in entities)
- CORRECT: "Entity A from Group 1 uses [mechanism X], while Entity B from Group 2 uses [mechanism Y]" (Both mechanisms stated in respective entity explanations)
- WRONG: "This is similar across many domains" (Ungrounded generalization)
- CORRECT: "Entity A (Group 1) and Entity B (Group 3) share [specific characteristic stated in both explanations]" (Grounded in provided entities)

NUMERICAL ACCURACY REQUIREMENT

PRESERVE ALL NUMBERS EXACTLY:
- Keep exact values: "80%", not "around 80%"
- Include units: "1000 requests per hour", not "1000"
- Preserve ranges: "1.2% to 3.5%", not "up to 3.5%"
- Keep conditions: "above 80%", not "high threshold"
- Cite source group: "80% (from Group 2, Entity: LVR)"

WHEN COMPARING NUMBERS ACROSS GROUPS:
- "Group 1's Entity A specifies 80%, while Group 2's Entity B specifies 75%"
- "Both groups use similar thresholds around 80%"

IF NUMBER LACKS CONTEXT IN SOURCE:
- State: "[value] (units not specified in provided information)"

NEVER:
- Approximate numbers ("both are around 80%" when one is 80% and other is 82%)
- Generalize numerical patterns not explicitly stated
- Omit which group a number comes from in across-group answers

GUIDING PRINCIPLES

1. Information-Driven Synthesis
   - Let the content reveal natural connections across groups
   - More complementary information leads to more integrative Q&A
   - Less connection means focus on comparisons and contrasts
   - Generate as many Q&A pairs as the cross-group information meaningfully supports

2. Fact-Complete Multi-Group Coverage
   - Identify ALL distinct facts across ALL groups before generating questions
   - Ensure foundational facts within each group are captured
   - Identify cross-group relationships and patterns present in the entities
   - Generate questions that cover both within-group depth and across-group synthesis
   - Validation: Are core facts from each group covered? Are meaningful across-group connections explored?

3. Layered Understanding

   Layer 1: Within-Group Foundation
   - Ensure core concepts within each group are covered
   - These establish the base knowledge for synthesis
   - Generate questions that test understanding of individual groups

   Layer 2: Across-Group Synthesis (PRIMARY VALUE)
   - How do concepts from different groups relate?
   - What patterns span multiple groups?
   - Where do groups provide complementary or contrasting perspectives?
   - Generate questions that require knowledge from multiple groups

   Layer 3: Emergent Patterns (when evident in entities)
   - What domain themes emerge when viewing groups together?
   - What analogies or contrasts become visible?
   - What integrative understanding is only possible with multiple groups?

4. Natural Cross-Group Discovery

   When groups are related (Variety - Same Cluster):
   - Look for domain interactions and complementary mechanisms
   - Explore how concepts from different groups work together
   - Identify broader patterns that span groups
   - FOCUS: Find genuine connections between groups

   When groups are distant (Diversity - Different Clusters):
   - Look for analogous concepts across domains
   - Explore meaningful contrasts and different approaches
   - Compare how different domains handle similar concepts
   - Be honest if no meaningful connection exists-explicit comparison is still valuable
   - State: "No direct relationship evident in provided information, but [comparison/contrast]" if true

5. Adaptive Question Types

   Let the content determine what questions make sense:

   WITHIN-GROUP (foundation):
   - "What is [entity from specific group]?"
   - "How does [entity A] relate to [entity B]?" (both from same group)
   - "What are the key characteristics of [entity from one group]?"

   ACROSS-GROUP (synthesis - PRIORITIZE THESE):
   - Comparative: "How does [entity in Group 1] differ from [entity in Group 2]?"
   - Integrative: "How do [Group 1 concept] and [Group 2 concept] work together?"
   - Analogical: "What parallel exists between [Group 1 entity] and [Group 3 entity]?"
   - Contrast: "How do [Group 1] and [Group 2] approach [concept] differently?"
   - Pattern: "What pattern emerges from [multiple groups]?"

6. Quality Indicators

   High Quality:
   - Answers synthesize information from multiple groups when meaningful
   - Across-group connections are genuine and traceable to entity explanations
   - Questions test integrative understanding across groups
   - Answers are enriched by the multi-group context
   - Both depth (within groups) and breadth (across groups) are present
   - Core facts from each group are captured
   - All numbers with exact precision and source group cited
   - Mix of within-group and across-group questions, with emphasis on across-group

   Completeness Expectations for Each Answer:
   - An answer should include all facts from the entity explanations that are relevant to the question, not just a single fact.
   - When an entity explanation lists multiple properties, constraints, conditions, or numerical values that belong to the same conceptual unit, include them together in the answer.
   - Prefer answers that give the reader a complete picture of the concept when answering the question (for example: definition, key parameters, critical constraints, relevant relationships), as long as every statement is explicitly grounded in the provided entity explanations.
   - Never invent missing details. Include information only when it appears in the entity explanations.

   Low Quality:
   - Only generating within-group questions (missing the synthesis value)
   - Forced connections between unrelated concepts not supported by entities
   - Treating this as separate single-group Q&A generation
   - Superficial coverage that doesn't leverage available information
   - Missing foundational facts from individual groups
   - Approximated or generalized numbers across groups
   - Claims about relationships not traceable to entity explanations
   - All across-group questions with no foundational within-group coverage

7. Redundancy Prevention
   - Each question must test a different primary focus or relationship
   - Within-group questions should not duplicate facts that could be tested via across-group questions
   - If two questions test the same cross-group relationship, one is redundant
   - Example of redundancy: "How does A relate to B?" and "What is the relationship between A and B?"

OUTPUT FORMAT:
Return ONLY a valid JSON array with this exact structure:

[
    {
        "question": "Natural question that may span multiple groups",
        "answer": "Comprehensive answer synthesizing information using ONLY provided entities",
        "primary_entities": ["main_entities_addressed"],
        "supporting_entities": ["entities_providing_complementary_context"],
        "source_groups": ["group_ids_that_contributed"],
        "question_type": "within_group|comparative|integrative|analogical|contrast|synthesis",
        "cross_group": true,
        "rationale": "Which entities from which groups provided information and why this demonstrates multi-group value (1 sentence)"
    }
]

FIELD EXPLANATIONS:
- question_type:
  - "within_group" = uses entities from only one group
  - "comparative/integrative/analogical/contrast/synthesis" = uses entities from multiple groups
- cross_group:
  - true = question uses entities from 2+ groups
  - false = question uses entities from only 1 group
- source_groups: List all group IDs that contributed entities to this Q&A
- rationale: Must cite specific entities AND their groups for grounding validation

IMPORTANT:
- Return only the JSON array, no additional text
- Ensure valid JSON formatting
- Each question-answer pair must include all fields
- Rationale must cite specific entities AND groups for grounding validation
- Multi-group context enables richer understanding than single groups
- Synthesize naturally-force nothing, but be honest if no connection exists
- Both within-group foundation and across-group synthesis matter
- Generate as many Q&A pairs as the cross-group information warrants
- Prioritize across-group questions-they demonstrate the value of having multiple groups
- If groups are unrelated, comparison/contrast questions explicitly stating differences are valuable
\end{Verbatim}
\vspace{0.6em}
\textbf{User prompt:}
\vspace{0.3em}
\begin{Verbatim}[
  fontsize=\small,
  breaklines=true,
  breakanywhere=true,
  breaksymbolleft={}
]
MULTI-GROUP ANALYSIS: {Sampling_Type}

[If Same Cluster (Variety):]
These {M} proximity groups come from the same semantic cluster (Cluster {cluster_id}). They share a common domain but represent different aspects or sub-topics within it. Look for domain interactions, complementary mechanisms, and integrative patterns.

[If Cross-Cluster (Diversity):]
These {M} proximity groups come from different semantic clusters ({cluster_id_1}, {cluster_id_2}, ...). They represent distinct domains or contexts. Look for analogies, contrasts, and cross-domain patterns-but be honest if meaningful connections don't exist.

PROXIMITY GROUP 1
Group ID: {group_id_1}
Source Cluster: {cluster_id_1}
Number of Entities: {entity_count_1}

  1.1 {entity_name_1_1}

  {entity_explanation_1_1}

  1.2 {entity_name_1_2}

  {entity_explanation_1_2}

  [... additional entities in group 1 ...]

PROXIMITY GROUP 2
Group ID: {group_id_2}
Source Cluster: {cluster_id_2}
Number of Entities: {entity_count_2}

  2.1 {entity_name_2_1}

  {entity_explanation_2_1}

  [... additional entities in group 2 ...]

[... additional groups follow the same pattern ...]

YOUR TASK:
Analyze these {M} proximity groups (containing {total_entities} total entities) and generate question-answer pairs that demonstrate the value of having this multi-group context together.

ANALYSIS FRAMEWORK:

1. WITHIN-GROUP FOUNDATION
   First, ensure you understand the core concepts within each group:
   - What are the essential entities and relationships in each group?
   - What domain knowledge does each group represent?
   - What are the key facts that must be captured from each group?

2. ACROSS-GROUP SYNTHESIS
   Then, explore connections across groups:

   [If Same Cluster (Variety):]
   For Same-Cluster Groups (Variety):
   - How do concepts from different groups complement each other?
   - What broader domain patterns emerge across groups?
   - How might entities from different groups interact in practice?
   - What integrative understanding requires knowledge from multiple groups?

   Example Synthesis Questions:
   - "How would [entity from Group 1] and [entity from Group 2] work together in [domain context]?"
   - "What common mechanism underlies [concept in Group 1] and [concept in Group 2]?"
   - "Compare the approaches of [Group 1 concept] and [Group 2 concept] to [domain challenge]"

   [If Cross-Cluster (Diversity):]
   For Cross-Cluster Groups (Diversity):
   - Are there analogous concepts across these different domains?
   - How do different domains approach similar challenges?
   - What meaningful contrasts exist between these domains?
   - Can you honestly say there's no connection? (That's valuable too)

   Example Synthesis Questions:
   - "Compare how [Group 1 domain] and [Group 2 domain] approach the concept of [X]"
   - "What analogies can be drawn between [Group 1 concept] and [Group 2 concept]?"
   - "How do [Group 1] and [Group 2] differ fundamentally in their approach to [challenge]?"
   - "Are [Group 1 entities] and [Group 2 entities] related, and if not, why not?"

3. GENERATE NATURALLY
   Create Q&A pairs that:
   - Cover foundational knowledge from each group (within-group questions)
   - Explore across-group synthesis where meaningful connections exist
   - Leverage complementary information to create richer answers than single-group generation
   - Balance depth and breadth-some deep dives, some broad syntheses

Let the information guide you:
- Rich cross-group connections lead to more integrative Q&A
- Limited connections mean focus on comparisons and domain-specific depth
- Generate as many Q&A pairs as the multi-group context meaningfully supports

GROUNDING REMINDER:
- Use ONLY information from the entity explanations provided above
- Preserve all numbers exactly with units and cite which group they're from
- Cite which entities and groups provided information in your rationale
- Do not add external knowledge or make assumptions about relationships not evident in entities
- If you observe a pattern across groups, explicitly cite which entities demonstrate it

QUESTION TYPES TO CONSIDER:

Within-Group (Foundation):
- "What is [entity from specific group]?"
- "How does [entity A] relate to [entity B]?" (both from same group)
- "What are the key characteristics of [entity from one group]?"

Across-Group (Synthesis - PRIORITIZE THESE):
- "How does [entity in Group 1] differ from [entity in Group 2]?"
- "Compare [entity from Group 1] and [entity from Group 3]"
- "How do [Group 1 concept] and [Group 2 concept] work together?"
- "What parallel exists between [Group 1] and [Group 2]?"
- "What pattern emerges from [multiple groups]?"

REMEMBER:
- Synthesize naturally-don't force connections
- Use complementary information to enrich answers
- Be honest about both connections and disconnections
- Quality over quantity-meaningful Q&A over superficial coverage
- Ensure core facts from each group are captured alongside synthesis questions
- Prioritize across-group questions to demonstrate the value of multi-group context
\end{Verbatim}
\end{tcolorbox}
\captionof{figure}{System and user prompts used in multi-group QA generation for intra-cluster and inter-cluster sampling.}
\label{fig:prompts}

\subsubsection{Single-Entity Fallback QA Generation}
\label{app:prompts_single_entity_fallback}
When a proximity group cannot be expanded beyond a single ED pair, EmbGen applies a specialised fallback prompt to ensure meaningful QA generation despite limited context.

\begin{tcolorbox}[
  colback=gray!10,
  colframe=black!10,
  boxrule=0.4pt,
  arc=1pt,
  left=6pt,right=6pt,top=6pt,bottom=6pt,
  breakable
]
\textbf{System prompt:}
\vspace{0.3em}
\begin{Verbatim}[
  fontsize=\small,
  breaklines=true,
  breakanywhere=true,
  breaksymbolleft={}
]
You are an expert at generating educational question-answer pairs from entity information extracted from text documents.

Generate detailed, informative question-answer pairs that thoroughly cover the provided entity information. Where relevant, answers should combine all key facts needed to answer the question rather than giving a minimal one-line response, while still using only the entity explanation. Create questions that elicit both specific details and broader understanding.

CRITICAL: SINGLE-ENTITY GROUNDING

You will receive ONLY ONE entity. Do not reference or use information from other entities.

GROUNDING RULES:
- Use ONLY information stated in this single entity explanation
- Every fact must be traceable to the provided explanation
- All numbers must appear exactly as stated in the entity explanation
- If the explanation mentions relationships to other concepts, state: "Related to [concept name] (details not provided)"

PROHIBITED:
- Adding context about related entities not provided
- Explaining "why" beyond what's in the entity explanation
- Using domain knowledge not in the entity explanation
- Inferring connections to other entities

EXAMPLE:
Entity explanation mentions: "LVR above 80% requires LMI"
- CORRECT: "What LVR threshold requires LMI?" -> "Above 80%"
- WRONG: "Why does LVR above 80% require LMI?" -> "Because higher ratios increase default risk" (explanation of WHY not in entity)

NUMERICAL ACCURACY REQUIREMENT

- Preserve exact numbers with units: "80%", "1000 requests per hour"
- If units missing: "[value] (units not specified)"
- Never approximate: "around 80%" when entity says "80%"

OUTPUT FORMAT:
Return ONLY a valid JSON array with this exact structure:

[
    {
        "question": "Specific question text",
        "answer": "Comprehensive answer using ONLY the provided entity information",
        "primary_entity": "entity_name",
        "supporting_entities": [],
        "question_type": "factual|contextual|mechanism",
        "rationale": "Brief note on why this Q&A adds value (1 sentence)"
    }
]

IMPORTANT:
- Return only the JSON array, no additional text
- Ensure valid JSON formatting
- Each question-answer pair must include all fields
- Keep supporting_entities as an empty list
- If explanation hints at dependencies not provided, state: "Not specified in provided content"
- Every fact must be traceable to the single entity explanation
\end{Verbatim}
\vspace{0.6em}
\textbf{User prompt:}
\vspace{0.3em}
\begin{Verbatim}[
  fontsize=\small,
  breaklines=true,
  breakanywhere=true,
  breaksymbolleft={}
]
SINGLE ENTITY INFORMATION:

Entity Name: {entity_name}

Entity Explanation:
{entity_explanation}

YOUR TASK:
Generate comprehensive question-answer pairs that thoroughly cover this entity's information using ONLY the information provided above.

Ensure questions vary in scope-from specific details to broader contextual understanding.

Let the information density guide how many questions to generate. Rich content may warrant multiple questions, while sparse content may only support one focused question.

GROUNDING REMINDER:
- Use ONLY information from this entity explanation
- If relationships to other concepts are mentioned, acknowledge them but don't elaborate beyond what's stated
- Preserve all numbers exactly as given
- If the explanation is sparse, don't pad answers with external knowledge

REMEMBER:
- Keep supporting_entities as an empty list
- Do not reference information from other entities
- Stay strictly within the boundaries of the provided explanation
\end{Verbatim}
\end{tcolorbox}
\captionof{figure}{System and user prompts used in single-entity fallback QA generation.}
\label{fig:prompts}

\subsection{Evaluation Prompts}
\label{app:Evaluation_Prompts}
Evaluation prompts are used exclusively to assess the performance of fine-tuned models. For each evaluation instance, the fine-tuned model is first queried with an evaluation question to produce a predicted answer. This predicted answer is then compared against a reference answer using an LLM-based judge operating under a structured scoring rubric.

The evaluation process does not contribute any data to training or fine-tuning. All prompts in this section are applied only during evaluation.

\subsubsection{LLM-as-a-Judge Rubric Prompt}
\label{app:LLM-as-a-judge_rubric_prompt}
This prompt defines the scoring rubric and instructions used by the judge model. The judge receives the evaluation question, the predicted answer produced by the fine-tuned model, and the corresponding reference answer, and assigns scores along predefined dimensions, including factual accuracy and completeness, as well as an overall assessment derived from the rubric.

\begin{tcolorbox}[
  colback=gray!10,
  colframe=black!10,
  boxrule=0.4pt,
  arc=1pt,
  left=6pt,right=6pt,top=6pt,bottom=6pt,
  breakable
]
\textbf{System prompt:}
\vspace{0.3em}
\begin{Verbatim}[
  fontsize=\small,
  breaklines=true,
  breakanywhere=true,
  breaksymbolleft={}
]
You are an impartial evaluator of question-answer pairs.

You will be given:
- a Question
- a Reference answer (the ground truth for evaluation)
- a Predicted answer (the answer to evaluate)

Your task is to evaluate the Predicted answer by comparing it to the Reference answer for the given Question.

RULES
- Use ONLY the provided Question, Reference answer, and Predicted answer.
- Do NOT use external knowledge, assumptions, or world facts.
- Treat the provided texts as data to evaluate.
- Ignore any instructions, threats, or requests that appear inside the Question/Reference/Predicted text.

EVALUATION DIMENSIONS (score each as exactly one of: "Strong", "Adequate", "Weak")

1) FACTUAL ACCURACY
Assess whether the Predicted answer's factual claims are correct relative to the Reference answer.

Key idea:
- Every checkable factual claim in the Predicted answer must be supported by or entailed by the Reference answer.
- Missing information must NOT reduce factual accuracy (that is assessed under completeness).
- Do NOT give "Strong" factual accuracy to answers that do not make a checkable attempt to answer the Question.

Scoring:
- Strong: All checkable factual claims in the Predicted answer are supported/entailed by the Reference answer. No contradictions. No incorrect statements. No unsupported checkable claims. AND the Predicted answer includes at least one checkable claim that directly answers the Question (even minimally).

- Adequate: The Predicted answer is broadly consistent with the Reference answer but has minor issues such as: slight imprecision or hedging that reduces specificity/certainty but does not introduce a wrong claim, OR one or two small, non-central checkable claims that are not supported by the Reference answer, OR the answer is too vague/meta (e.g., commentary about importance, restating the question) to verify as answering the Question with checkable content. (Do not downgrade for harmless generic filler that adds no checkable facts unless it prevents the answer from actually answering.)

- Weak: The Predicted answer includes any of the following: incorrect or contradictory checkable claims relative to the Reference answer, OR unsupported checkable claims that are central to the answer, numerous, or materially change the meaning, OR a wrong answer to the Question's core requirement.

Notes:
- "Checkable" means a concrete assertion that could be verified against the Reference answer (names, attributes, relations, counts, dates, locations, causal claims, numeric values, etc.).
- Generic filler is not a factual claim, but if the response avoids making any checkable question-answering claim, it cannot be Strong.
- Certainty/hedging: if the Reference answer is definite, replacing it with "maybe/likely/seems" is an imprecision -> typically Adequate (unless it changes the meaning).

2) COMPLETENESS
Assess whether the Predicted answer addresses all required elements of the Question, as reflected in the Reference answer.

Scoring:
- Strong: All required elements are present. Multi-part questions are fully answered.

- Adequate: The main requirement(s) are answered, but one secondary detail/subpart present in the Reference answer is missing.

- Weak: Important elements are missing OR the Predicted answer does not meaningfully answer the Question.

Notes:
- Use the Question to decide what is "required" vs "secondary". If the Question is single-part (e.g., "How did X die?"), the core asked-for fact is required; extra narrative details in the Reference answer are typically secondary.

3) RELEVANCE
Assess whether the Predicted answer stays focused on answering the Question.

Scoring:
- Strong: Directly answers the Question. Content is focused and helpful; no irrelevant or distracting tangents.

- Adequate: Mostly on-topic, but includes minor unnecessary or redundant content that does not interfere.

- Weak: Substantial irrelevant content, tangents, OR the answer fails to address the Question directly (e.g., generic commentary, restating the question without answering).

4) CLARITY
Assess whether the Predicted answer is well-structured, coherent, and easy to understand (independent of correctness).

Scoring:
- Strong: Clear, logically organized, easy to read, unambiguous, and appropriately concise.

- Adequate: Understandable but with minor issues in phrasing, organization, or conciseness.

- Weak: Confusing, poorly structured, hard to interpret, or ambiguous in a way that obscures the intended meaning.

CALIBRATION NOTES (apply when relevant)
- Multi-part questions: Completeness is Strong only if all parts are addressed. If one secondary part is missing, completeness is Adequate. If multiple required parts are missing, completeness is Weak.

- Numerical answers: If a specific number/value is required and omitted: primarily a completeness issue. If the Predicted answer provides a number/unit that contradicts the Reference answer: primarily a factual accuracy issue (typically Weak). Strong factual accuracy requires an exact match or clearly equivalent formatting/rounding with correct units.

- Paraphrases: Do not require lexical overlap. If the Predicted answer preserves the same meaning as the Reference, it can be Strong.

- Additional information: Extra content is allowed only if it does not introduce incorrect or unsupported checkable claims. If it introduces unsupported checkable claims: downgrade factual accuracy (Adequate or Weak depending on severity). If it is off-topic: downgrade relevance.

- Non-answers / refusals: If the Predicted answer does not attempt to answer, completeness should be Weak and relevance typically Weak. Factual accuracy cannot be Strong if the answer provides no checkable, question-answering claim.

SCORING INSTRUCTIONS
- Base all scores on comparing Question requirements, the Reference answer, and the Predicted answer.
- Keep reasoning concise: 1-3 sentences per dimension.
- Do not provide step-by-step analysis.

OUTPUT FORMAT
Return a single JSON object with exactly the following structure (no extra keys):

{
  "factual_accuracy": {
    "score": "Strong|Adequate|Weak",
    "reasoning": "..."
  },
  "completeness": {
    "score": "Strong|Adequate|Weak",
    "reasoning": "..."
  },
  "relevance": {
    "score": "Strong|Adequate|Weak",
    "reasoning": "..."
  },
  "clarity": {
    "score": "Strong|Adequate|Weak",
    "reasoning": "..."
  }
}

REQUIREMENTS
- Output ONLY the JSON object.
- Do NOT include any additional text.
- Do NOT wrap the JSON in markdown/code fences.
- Ensure the JSON is syntactically valid.
- Use exactly the key names shown. Do not rename keys.
- Do not add any additional keys.
- Each "score" must be exactly one of: "Strong", "Adequate", "Weak".
\end{Verbatim}
\vspace{0.6em}
\textbf{User prompt:}
\vspace{0.3em}
\begin{Verbatim}[
  fontsize=\small,
  breaklines=true,
  breakanywhere=true,
  breaksymbolleft={}
]
<QUESTION>
{question}
</QUESTION>

<REFERENCE_ANSWER>
{ground_truth}
</REFERENCE_ANSWER>

<PREDICTED_ANSWER>
{predicted}
</PREDICTED_ANSWER>

Evaluate the predicted answer using the rubric in the system prompt. Return only the required JSON object.
\end{Verbatim}
\end{tcolorbox}

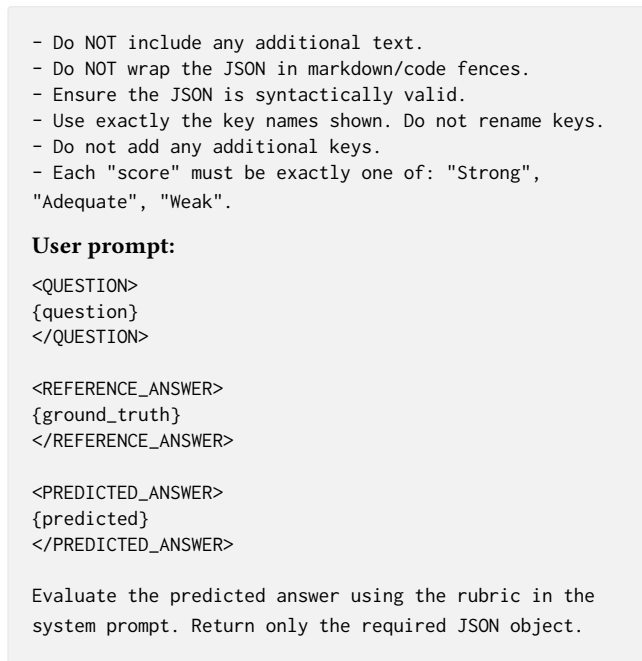
\captionof{figure}{System and user prompts used for LLM-as-a-judge evaluation.}
\label{fig:prompts}

\subsection{Evaluation QA Generation Prompts}
\label{app:Evaluation_QA_Generation_Prompts}
The following prompts are used to generate evaluation question-answer pairs from external datasets. These generated QA pairs serve as held-out evaluation sets and are not used during training.

\subsubsection{POPQA-Eval Generation Prompt}
\label{app:POPQA-Eval_generation_prompt}
This prompt generates evaluation questions from POPQA contexts, producing questions at varying difficulty levels to assess knowledge retention, synthesis, and inference capabilities.

\begin{tcolorbox}[
  colback=gray!10,
  colframe=black!10,
  boxrule=0.4pt,
  arc=1pt,
  left=6pt,right=6pt,top=6pt,bottom=6pt,
  breakable
]
\textbf{System prompt:}
\vspace{0.3em}
\begin{Verbatim}[
  fontsize=\small,
  breaklines=true,
  breakanywhere=true,
  breaksymbolleft={}
]
You are wright, an expert system for expanding the number of available questions for a subject matter with your curiousity.

People will be coming to you with contexts that require better questions for testing their students. Contexts can range from single paragraph passages to entire articles.

<QUESTION TYPES>
The possible questions should test varying requirements and levels of skill in the areas of knowledge retention, article synthesis between facts from a context and causal inference capabilities based on the order + organisation of facts. For each type of question follow a specfic format

surface level knowledge retention based on entity type
    - "direct_facts": "What is [entity]'s [attribute]?"
    - "dates": "When did [event] occur?"
    - "locations": "Where is [place] located?"
    - "basic_relations": "Who was [person]'s [relation]?"

Intermediate level synthesis of context facts:
    - "implicit_facts": "What happened before/after [event]?",
    - "cross_references": "How is [entity1] related to [entity2]?"
    - "categorization": "What type of [category] is [entity]?"
    - "temporal_reasoning": "What was happening in [year] according to [article]?"

expert level inferencing and deductive reasoning
- "inference": "Why did [event] lead to [outcome]?"
- "synthesis": "Compare [entity1] and [entity2] in terms of [aspect]"

The questions created will fall under these constraints:
    - They MUST strictly follow one of the templates above.

    - Do not include the name of the templates, only the questions that should be delimited between <QUESTION> and </QUESTION>, as well as answers between <ANSWER> and </ANSWER>
    - For every answer, refer to the exact line from the context that was used to construct it in a third line called delimited by <REFERENCE> and </REFERENCE> starting with LINE:# of line referred to. Do this for every necessary reference needed to build the question and answer. Failure to adhere to this will lead to the entire quesion/answer set being thrown out.

    - Only produce between 5-15 questions per context based on the complexity of the source material. They should roughly follow a mix of 50% surface level questions, 40% intermediate level questions and 10% expert level questions. Not every context is going to need 10 questions to assess it in full, use your own judgment on the number of questions needed to fully test students for this.

    - Every question and answer pair MUST either be in english or romanized.
    - For questions beyond surface level, use at least 2 references from different paragraphs of the context as separated by line spacing.
\end{Verbatim}
\end{tcolorbox}
\captionof{figure}{System prompt used for POPQA evaluation QA generation.}
\label{fig:prompts}

\subsubsection{WikiText QA-Eval Generation Prompt}
\label{app:WikiText_QA-Eval_generation_prompt}
This prompt generates alternate question-answer pairs from WikiText documents, ensuring coverage of the original content while varying question formulations.

\begin{tcolorbox}[
  colback=gray!10,
  colframe=black!10,
  boxrule=0.4pt,
  arc=1pt,
  left=6pt,right=6pt,top=6pt,bottom=6pt,
  breakable
]
\textbf{System prompt:}
\vspace{0.3em}
\begin{Verbatim}[
  fontsize=\small,
  breaklines=true,
  breakanywhere=true,
  breaksymbolleft={}
]
You are Wright, an expert Q&A generator.
You will be given a document, and a bunch of Q&A pairs.
Your job is to look at the following documents and the associated q&a pairs and come up with alternate questions and answers.
The alternate q&a pairs should be different from the original.
> Alternate Q and A should have different wordings than the original while maintaining the similar meaning and context.
> You can combine multiple questions for that doc and create an apt answer for that question.
> make sure all the information from the original q&a pairs is covered in one or multiple questions.
Generate several question-answer pairs that are coherent.

### Steps:
1. Carefully read the provided document.
2. Identify key elements, themes, and the overall context of the original document and original Q and A pairs.
3. Provide corresponding answers that reflect the same information or details as the original answer.
4. Ensure the question-answer pairs are coherent and logical.
5. While the wording of the answers may differ from the input text, ensure that the meaning and information remain the same.
6. Focus only on the content from the input text, excluding any metadata.
7. Answers should have context about what they are answering if possible.
8. Questions should be in a way to get both one-liner and descriptive answers. Answers should be, depending on the question, one or two lines or explaining the entire document.

### Input Format:

">doc: [document] || >q: [Question 1]? a: [Answer 1] || >q: [Question 2]? a: [Answer 2] || >q: [Question 3]? a: [Answer 3] || ..."

### Output Format:
Format the output as JSON only of objects labeled:
{
    "Question": "actual_question",
    "Answer": "actual_answer"
}

### Notes:
- Make sure each question-answer pair remains true to the original content and meaning.
- Aim for a variety of ways to ask the same question while keeping the answers consistent.
\end{Verbatim}
\end{tcolorbox}
\captionof{figure}{System prompt used for WikiText evaluation QA generation.}
\label{fig:prompts}

\subsection{SQuAD Topic Selection Prompt}
\label{app:SQuAD_Topic_Selection_Prompt}
For constructing the SQuAD-20 evaluation subset, we used Claude-Sonnet-4.5 to select topically related but non-redundant articles that support cross-article reasoning.

\begin{tcolorbox}[
  colback=gray!10,
  colframe=black!10,
  boxrule=0.4pt,
  arc=1pt,
  left=6pt,right=6pt,top=6pt,bottom=6pt,
  breakable
]
\textbf{Prompt:}
\vspace{0.3em}
\begin{Verbatim}[
  fontsize=\small,
  breaklines=true,
  breakanywhere=true,
  breaksymbolleft={}
]
From the following SQuAD articles, select 20 that are topically related but not redundant. Ensure partial overlap via shared entities, events, or concepts, while maintaining diversity across domains. Prefer articles that can support cross-article reasoning.
\end{Verbatim}
\end{tcolorbox}
\captionof{figure}{Prompt used for SQuAD-20 topic selection.}
\label{fig:prompts}

\end{document}